\documentclass[journal]{IEEEtran}
\usepackage{graphicx}
\usepackage{subfigure}
\usepackage{algorithmicx}
\usepackage[ruled]{algorithm}
\usepackage{algpseudocode}
\usepackage{amsmath}
\usepackage{times}
\usepackage{color}
\usepackage{amsthm}
\usepackage{epstopdf}

\ifCLASSINFOpdf
\else
\fi

\begin{document}
%
\title{F-SVM: Combination of Feature Transformation and SVM Learning via Convex Relaxation}

\author{Xiaohe Wu,
        Wangmeng Zuo,~\IEEEmembership{Senior Member,~IEEE,}
        Yuanyuan Zhu,
        Liang Lin,~\IEEEmembership{Member,~IEEE}
\thanks{X. Wu, W. Zuo and Y. Zhu are with School of Computer Science and Technology, Harbin Institute of Technology, Harbin, China.}%
\thanks{L. Lin is with School of Advanced Computing, Sun Yat-sen University, Guangzhou, China.}%
}

\maketitle

\begin{abstract}
The generalization error bound of support vector machine (SVM) depends on the ratio of radius and margin, while standard SVM only considers the maximization of the margin but ignores the minimization of the radius. Several approaches have been proposed to integrate radius and margin for joint learning of feature transformation and SVM classifier. However, most of them either require the form of the transformation matrix to be diagonal, or are non-convex and computationally expensive.

In this paper, we suggest a novel approximation for the radius of minimum enclosing ball (MEB) in feature space, and then propose a convex radius-margin based SVM model for joint learning of feature transformation and SVM classifier, i.e., F-SVM. An alternating minimization method is adopted to solve the F-SVM model, where the feature transformation is updated via gradient descent and the classifier is updated by employing the existing SVM solver. By incorporating with kernel principal component analysis, F-SVM is further extended for joint learning of nonlinear transformation and classifier. Experimental results on the UCI machine learning datasets and the LFW face datasets show that F-SVM outperforms the standard SVM and the existing radius-margin based SVMs, e.g., RMM, R-SVM$^{+}$ and R-SVM$_{\mu}^{+}$.

\end{abstract}

\begin{IEEEkeywords}
Support vector machine, radius margin bound, convex relaxation, max-margin.
\end{IEEEkeywords}

%
\IEEEpeerreviewmaketitle

\section{Introduction}
\IEEEPARstart{S}{upport} vector machine (SVM) and its extensions have been one of the most successful machine learning methods \cite{vapnik1998statistical,cristianini2000introduction}, and have been adopted in various fields, e.g., computer vision \cite{osuna1997training, heisele2001face, guo2001support, wu2015linear}, signal processing \cite{samanta2003artificial, chen2001support}, natural language processing \cite{collobert2011natural, tsochantaridis2004support} and bioinformatics \cite{do2009feature, weston2000feature, guyon2002gene, rakotomamonjy2003variable}. Despite its popularity, SVM aims to seek the optimal hyperplane with the maximum margin principle, but the generalization error of SVM actually is a function of the ratio of radius and margin \cite{vapnik2000bounds}. Given feature space, the radius is fixed and can be ignored, thus SVM can minimize the generalization error by maximizing the margin. However, for joint learning of feature transformation and classifier, the radius information will be valuable and cannot be ignored.

By minimizing the radius-margin ratio, the generalization error of SVM can be optimized for joint learning of feature transformation and classifier. Since the radius-margin error bound is non-convex, relaxation and approximation of radius is generally adopted in the existing models \cite{shivaswamy2010maximum, zhu2012learning}. Several approaches have been proposed from the perspective of radius-margin error \cite{do2009feature, shivaswamy2010maximum, zhu2012learning, do2013convex}, but most ones suffer from the limitations of computational burden and simplified forms of transformation. RMM \cite{shivaswamy2010maximum} only considers the spread of the data along the direction perpendicular to the classification hyperplane. Radius-margin based SVMs, e.g., MR-SVM \cite{do2009feature}, R-SVM$^{+}$ \cite{do2013convex} and RSVM$_{\mu}^{+}$ \cite{do2013convex}, are based on the constraint that the linear transformation matrix should be diagonal.

Another strategy for joint feature transformation and classifier learning is to incorporate metric learning with SVM, where metric learning can be adopted to learn a better linear transformation matrix \cite{xing2002distance, zhu2012learning, globerson2005metric, schultz2004learning, wang2013kernel, shen2014efficient, guillaumin2009you, wang2014shrinkage, tran2008human}. One simple approach to combine metric learning and SVM is to directly deploy the transformation obtained using metric learning into SVM. This approach, however, usually cannot lead to satisfactory performance improvement \cite{xu2012distance}. Therefore, other approaches have been proposed to integrate metric learning to SVM, e.g., support vector metric learning (SVML) \cite{xu2012distance} and metric learning with SVM (MSVM) \cite{zhu2012learning}. But SVML \cite{xu2012distance} was designed for RBF-SVM and ignored the radius information, while MSVM \cite{zhu2012learning} is non-convex.

In this paper, we propose a novel radius-margin based SVM model for joint learning of feature transformation and SVM classifier, i.e., F-SVM. Compared with the existing radius-margin based SVM methods, we derive novel lower and upper bounds for the relaxation of the radius-margin ratio. Unlike MR-SVM \cite{do2009feature}, R-SVM$^{+}$ \cite{do2013convex} and RSVM$_{\mu}^{+}$ \cite{do2013convex} which are suggested for joint feature weighting and SVM learning, F-SVM can simultaneously learn feature transformation ${\bf L}$ and classifier $( {\bf w}, b )$. Compared with the existing metric learning for SVM methods, our F-SVM model considers both the radius and the margin information, and is convex. Then, an alternating minimization algorithm is proposed to solve our F-SVM model, which iterates by updating feature transformation and classifier alternatively. Note that kernel SVM is equivalent to perform linear SVM in the kernel PCA space. We further suggest to conduct linear FSVM in the kernel PCA space for joint learning of nonlinear transformation and classifier. The contribution of this paper is of three-fold:
\begin{itemize}

    \item A novel convex formulation of radius-margin based SVM model, i.e., F-SVM, is proposed. Unlike MR-SVM \cite{do2009feature}, R-SVM$^{+}$ \cite{do2013convex} and RSVM$_{\mu}^{+}$ \cite{do2013convex}, our F-SVM is capable of joint learning feature transformation and classifier, and is robust against outliers. Experimental results show that F-SVM outperforms SVM and the existing radius-margin based SVMs.

    \item  In F-SVM, we derive the lower and upper bounds for the radius of minimum enclosing ball (MEB) in feature space, resulting in a novel approximation of the radius. Compared with the approximations proposed in \cite{do2013convex}, ours is much simple and can be easily adopted in developing radius-margin based SVM models.

    \item An alternating minimization algorithm is developed for solving F-SVM via iterating between gradient descent and SVM learning. Therefore, the off-the-shelf SVM solvers can be employed to improve the computational efficiency. Moreover, a semi-whitened PCA method is developed for the initialization of $ {\bf M} = {\bf L}^T {\bf L} $.

\end{itemize}

The remainder of the paper is organized as follows: Section 2 reviews the related work on the radius-margin ratio based bounds and their applications. Section 3 describes the model and algorithm of the proposed F-SVM method. Section 4 extends F-SVM to the kernelized version for nonlinear classification. Section 5 provides the experimental results on the UCI machine learning datasets and the LFW dataset. Finally, we conclude the paper in Section 6.

\section{Related work}
The radius-margin bound not only provides theoretical explanation on the generalization performance of SVM \cite{vapnik1998statistical}, but also has been extensively adopted for improving kernel classification methods, e.g., model selection \cite{chapelle2002choosing, sydorov2014deep}, multiple kernel learning (MKL) \cite{gai2010learning, do2009margin, liu2013efficient, liu2014efficient}, and mapping of nominal attributes \cite{peng2014improved}. Denote a training set by $\mathcal{S}=\left\{ \left( {{\mathbf{x}}_{1}},{{y}_{1}} \right),...,\left( {{\mathbf{x}}_{n}},{{y}_{n}} \right) \right\}$ and a feature space by $\mathcal{H} : \Phi \left( \mathbf{x} \right)$. In \cite{gai2010learning, do2009margin}, the radius ${R}$ of minimum enclosing ball (MEB) in feature space is computed as:
\begin{eqnarray} \label{eq:class_R}
\min_{ {R}, {\mathbf{x}}_{0} } {R} ^{2}, s.t. \| \Phi ({{\mathbf{x}}_{i}})-\Phi ({{\mathbf{x}}_{0}}) \|_{2}^{2} \le {{R}^{2}},i=1,2,\cdots ,n.
\end{eqnarray}
Assuming that the training set is separable, given the optimal hyperplane $( {\bf w}, b )$, Vapnik \cite{vapnik1998statistical} suggested a radius-margin error bound which showed that the expectation of the misclassification probability depends on ${R}^{2} \| {\mathbf{w}} \|_{2}^{2}$.

The standard SVM is known as a max-margin method which only considers the margin $1/\| {\mathbf{w}} \|_{2}^{2}$  in the algorithm. When the feature space is fixed, the radius is a constant and can thus be ignored. But in many classification tasks, the model parameters \cite{chapelle2002choosing}, combination of basis kernels \cite{do2009margin}, feature reweighting or transformation \cite{weston2000feature} usually should be learned or tuned based on the training data, where integration of radius has been demonstrated to be very effective in improving the classification performance. In model selection, radius-margin bound has been applied for choosing tradeoff parameter and scaling factors of SVM and $L_{1}$-SVM \cite{chapelle2002choosing}. In multiple kernel learning (MKL) \cite{gai2010learning, do2009margin, liu2013efficient} and feature reweighting \cite{weston2000feature}, several variants of radius had been developed.

This paper aims to jointly learn SVM together with feature transformation by minimizing the radius-margin ratio, i.e., radius-margin based SVM, and more detailed review is given on this topic. Except \cite{zhu2012learning}, most existing approaches \cite{do2009feature, do2013convex, schultz2004learning} require the transformation matrix to be diagonal, i.e., feature reweighting and selection. Direct use of radius-margin ratio ${R}^{2} \| {\mathbf{w}} \|_{2}^{2}$ in SVM results in a non-convex optimization problem, which makes the learning algorithm computationally expensive and unstable. By restricting the feature transformation to be diagonal ${{\mathbf{D}}_{\bf{\mu }}}\!=\!Diag\left( \bf{\mu } \right)$ with ${{\mu }_{k}} \!\ge\! 0$, Do et al. \cite{do2009feature} suggested that the radius is bounded with ${{\max }_{k}}\ {{\mu }_{k}}R_{k}^{2}\le R_{\mathbf{\mu }}^{2}\le \sum\nolimits_{k}{{{\mu }_{k}}R_{k}^{2}}$, where ${{R}_{k}}$ is the radius on dimension ${k}$. By approximating $R_{\mathbf{\mu }}^{2}$ with its upper bound $\sum\nolimits_{k}{{{\mu }_{k}}R_{k}^{2}}$, MR-SVM in \cite{do2009feature} solved the following convex relaxation problem:
\begin{eqnarray} \label{eq:class_MR-SVM}
\begin{aligned}
 \min \limits_{ {\bf w},{b},{\mathbf \xi},{\mathbf \mu} } \ \ \ \ & {\frac{1}{2}}\sum\nolimits_{k}\frac{w_{k}^{2}}{{\mu}_{k}}+{\frac{C}{\sum\nolimits_{k}{{{\mu }_{k}}R_{k}^{2}}}}\sum\nolimits_{i}{{{\xi }_{i}}}, \\
 s.t. \ \ \ \ & {{y}_{i}}({{{\mathbf{w}}^{T}}{{\mathbf{x}}_{i}}}+b)\ge 1-{{\xi }_{i}}, \forall i, \\
 \ \ \ \ & {{\xi }_{i}} \ge 0, i=1,2,\cdots ,n,\\
 \ \ \ \ & \sum\nolimits_{k}{{{\mu }_{k}}}=1,{{\mu }_{k}}\ge 0,\forall k.
\end{aligned}
\end{eqnarray}
Denote ${{R}_{O}}$ by the half value of the maximum pairwise distances. Do et al. in \cite{do2013convex} introduced a tighter bound of the radius ${{R}_{O}}\le {{R}_{\mathbf{\mu }}}\le \frac{1+{\sqrt{3}}}{2}{{R}_{O}}$ and proposed another convex model R-SVM$_{\mu}^{+}$:
\begin{eqnarray} \label{eq:class_R-SVM-mu+}
\begin{aligned}
 \min \limits_{ {\bf w},{b},{\mathbf \xi},{\mathbf \mu},{r}} \ \ \ \ & {\frac{1}{2}}\sum\nolimits_{k}\frac{w_{k}^{2}}{{\mu}_{k}}+{\lambda}{r}+ C\sum\nolimits_{i}{{{\xi }_{i}}}, \\
 s.t. \ \ \ \ & {{y}_{i}}({{\mathbf{w}}^{T}}{{\mathbf{x}}_{i}}+b)\ge 1-{{\xi }_{i}}, \forall i, \\
  \ \ \ \ & {{\xi }_{i}} \ge 0, i=1,2,\cdots ,n, \\
 \ \ \ \ & \sum\nolimits_{k}{{{\mu }_{k}}}=1,{{\mu }_{k}}\ge 0,\forall k, \\
 \ \ \ \ & \frac{1}{2}{{\left\| {{D}_{\sqrt{\mathbf{\mu }}}}{{\mathbf{x}}_{i}}-{{D}_{\sqrt{\mathbf{\mu }}}}{{\mathbf{x}}_{j}} \right\|}^{2}}\le r,\forall i,j.
\end{aligned}
\end{eqnarray}
Furthermore, R-SVM$^{+}$ was developed in \cite{do2013convex} by controlling both the radius and margin with ${\bf w}$.

Rather than feature reweighting and selection, Zhu et al. \cite{zhu2012learning} proposed a metric learning with SVM (MSVM) method for joint learning of the linear transformation and SVM classifier. In \cite{zhu2012learning}, given the transformation matrix ${\mathbf A}$, an alternative $\bar{R}={{\max }_{i}}\left\| \mathbf{A}{{\mathbf{x}}_{i}}-\mathbf{A\bar{x}} \right\|_{2}^{2}$ of the radius $R$ was adopted, where $\mathbf{\bar{x}}$ is the mean of the training samples. Although Zhu et al. \cite{zhu2012learning} claimed that $R=\bar{R}$, as demonstrated in {\textbf {Theorem 1}} of this work, $\bar{R}$ is an upper bound of $R$. The MSVM model in \cite{zhu2012learning} was formulated as:
\begin{eqnarray} \label{eq:class_MSVM}
\begin{aligned}
 \min \limits_{\mathbf{w},b,\mathbf{A}} \ \ \ \ & \frac{1}{2}\left\| \mathbf{w} \right\|_{2}^{2}+C\sum\nolimits_{i}{{{\xi }_{i}}}, \\
 s.t. \ \ \ \ & {{y}_{i}}({{\mathbf{w}}^{T}}\mathbf{A}{{\mathbf{x}}_{i}}+b)\ge 1-{{\xi }_{i}}, \forall i, \\
  \ \ \ \ & {{\xi }_{i}} \ge 0, i=1,2,\cdots ,n, \\
 \ \ \ \ & {{\left\| \mathbf{A}{{\mathbf{x}}_{i}}-\mathbf{A\bar{x}} \right\|}^{2}}\le 1,\forall i.
\end{aligned}
\end{eqnarray}
Note that MSVM is non-convex and solved using gradient projection.

In this paper, we propose a novel relaxed convex model of radius-margin based SVM, i.e., F-SVM, for joint learning of feature transformation and SVM classifier. Compared with the existing radius-margin based SVM methods, F-SVM has some distinguishing features. Our F-SVM model is convex, while MSVM \cite{zhu2012learning} is non-convex. Unlike RMM \cite{shivaswamy2010maximum}, the transformation in F-SVM is learned to minimize the radius of the enclosing ball of all samples rather than to only shrink the sample span along the direction perpendicular to the hyperplane. Moreover, F-SVM is also different with MR-SVM \cite{do2009feature}, R-SVM$^{+}$ \cite{do2013convex} and R-SVM$_{\mu}^{+}$ \cite{do2013convex} from three aspects: $(i)$ Instead of feature reweighting and selection, F-SVM can learn feature transformation and classifier simultaneously; $(ii)$ F-SVM adopts a new approximation for the radius of MEB in feature space; $(iii)$ In F-SVM, individual inequality constraints are combined into one holistic inequality constraint to improve the robustness against outliers. All these make F-SVM very promising for joint learning of feature transformation and SVM classifier, and the results also validate the effectiveness of F-SVM.

\section{Radius-margin based Support Vector Machine}

\subsection{Problem Formulation}
Denote $\mathcal{S}=\left\{ \left( {{\mathbf{x}}_{1}},{{y}_{1}} \right),...,\left( {{\mathbf{x}}_{n}},{{y}_{n}} \right) \right\}$ by a training set, where ${{\mathbf{x}}_{i}} \in {{\mathbf{R}}^{d}}$ and ${{y}_{i}}\in \left\{ -1,+1 \right\}$ denote the {\it i}th training sample and the corresponding class label, respectively. By introducing the slack variables ${{\xi }_{i}}\left( i=1,2,...,n \right)$, SVM aims to find the optimal separating hyperplane by solving the following optimization problem:
\begin{eqnarray} \label{eq:class_SVM}
\begin{aligned}
 \min \limits_{\mathbf{u},b,{\xi}} \ \ \ \ & \frac{1}{2}\left\| \mathbf{u} \right\|_{2}^{2}+C\sum\nolimits_{i}{{{\xi }_{i}}}, \\
 s.t. \ \ \ \ & {{y}_{i}}({{\mathbf{u}}^{T}}{{\mathbf{x}}_{i}}+b)\ge 1-{{\xi }_{i}}, \forall i, \\
 \ \ \ \ & {{\xi }_{i}} \ge 0, i=1,2,\cdots ,n.
\end{aligned}
\end{eqnarray}
where $({\bf u},b)$ are the parameters to describe the learned hyperplane ${{\mathbf{u}}^{T}}\mathbf{x}+b=0$, ${\xi}_{i}$ denotes the {\it i}th slack variable, and $C$ stands for the tradeoff parameter. The objective function in Eq. (\ref{eq:class_SVM}) aims to maximize the margin $\gamma ={1}/{{{\left\| \mathbf{u} \right\|}^{2}}}$ while minimizing the empirical risk $\sum\nolimits_{i=1}^{n}{{{\xi }_{i}}}$. For joint learning, we introduce a linear transformation matrix ${\bf A}$ and integrate the radius information, resulting in the following radius-margin based SVM model:
\begin{eqnarray} \label{eq:class_radiu-margin-based-SVM}
\begin{aligned}
 \min \limits_{\mathbf{u},b,{\xi},{\mathbf{A}},R} \ \ \ \ & \frac{1}{2}\left\| \mathbf{u} \right\|_{2}^{2}{R}^2+C\sum\nolimits_{i}{{{\xi }_{i}}}, \\
 s.t. \ \ \ \ & {{y}_{i}}({{\mathbf{u}}^{T}}{\bf A}{{\mathbf{x}}_{i}}+b)\ge 1-{{\xi }_{i}}, \forall i, \\
 \ \ \ \ & {{\xi }_{i}} \ge 0, i=1,2,\cdots ,n.
\end{aligned}
\end{eqnarray}
where the radius $R$ is defined as:
\begin{eqnarray} \label{eq:class_F-SVM-R}
\min_{ {R}, {\mathbf{x}}_{0} } {R} ^{2}, s.t. \| {{\bf A}{\mathbf{x}}_{i}}- {{\bf A}{\mathbf{x}}_{0}} \|_{2}^{2} \le {{R}^{2}},i=1,2,\cdots ,n.
\end{eqnarray}
Note that $R^2$ depends on matrix $\bf A$ and the problem in Eq. (\ref{eq:class_radiu-margin-based-SVM}) is non-convex \cite{do2013convex}. Denote ${\bf {x}_{0}}$ by the center of all instances, and $\bar R$ by the largest squared distance from the center in transformed feature space. Let ${{\mathbf{x}}_{0}}=\mathbf{\bar{x}}=\sum\nolimits_{i=1}^{n}{{{\mathbf{x}}_{i}}}$ and $\bar{R}= \max_{i} \left\| {\bf A}{{\bf x}_{i}}-{\bf A}{\bf \bar{x}} \right\|_{2}^{2}$. We prove that the radius $R$ is bounded by $\bar R$.

{\textbf{Theorem 1}}. The radius $R$ is bounded by $\bar R$ by:
\begin{eqnarray} \label{eq:class_R-bound}
\frac{1}{2}\overline{R}\le R\le \overline{R}.
\end{eqnarray}
Please refer to {\textbf{Appendix A}} for the proof of {\textbf{Theorem 1}}. In \cite{zhu2012learning}, Zhu et al. claimed that $R=\bar{R}$ . From {\textbf{Theorem 1}}, $\bar R$ is only an approximation of $R$, and counter examples can be easily found to illustrate $R\ne \bar{R}$. Let $\mathbf{w}={{\mathbf{A}}^{T}}\mathbf{u}$ and $\mathbf{M}={{\mathbf{A}}^{T}}\mathbf{A}$. Since the radius $R$ is upper bounded by $\bar R$, we can approximate $R$ with $\bar R$. With simple algebra, the radius-margin SVM model in Eq. (\ref{eq:class_radiu-margin-based-SVM}) is relaxed into the following formulation:
\begin{eqnarray} \label{eq:class_radiu-margin-based-SVM-relaxation}
\begin{aligned}
 \min_{\mathbf{w}\!,b\!,\mathbf{\xi }\!,\mathbf{M}\!,\bar{R}\!} \ \ \ \ & F(\mathbf{w},b,\mathbf{\xi },\mathbf{M},\bar{R}) \!\!=\!\! \frac{1}{2}\left( {{\mathbf{w}}^{T}}\!{{\mathbf{M}}^{-1}}\!\mathbf{w}\! \right)\bar{R}_{{}}^{2}\!+\!C\!\sum\limits_{i=1}^{n}{{{\xi }_{i}}}, \\
 s.t. \ \ \ \ &  {{y}_{i}}({{\mathbf{w}}^{T}}{{\mathbf{x}}_{i}}+b)\ge 1-{{\xi }_{i}}, \forall i,\\
 \ \ \ \ & {{\xi }_{i}}\ge 0,i=1,2,\cdots ,n, \\
 \ \ \ \ & {{({{\mathbf{x}}_{i}}-\mathbf{\bar{x}})}^{T}}\mathbf{M}({{\mathbf{x}}_{i}}-\mathbf{\bar{x}})\le {{{\bar{R}}}^{2}}.
\end{aligned}
\end{eqnarray}

{\textbf{Theorem 2}}. The problem in Eq. (\ref{eq:class_radiu-margin-based-SVM-relaxation}) is equivalent with the following problem:
\begin{eqnarray} \label{eq:class_radiu-margin-based-SVM-relaxation-equivalentaion}
\begin{aligned}
 \min_{\mathbf{w},b,\mathbf{\xi },\mathbf{M}} \ \ \ \ & L(\mathbf{w},b,\mathbf{\xi },\mathbf{M})\!\!=\!\!
\left\{ \frac{1}{2}\left( {{\mathbf{w}}^{T}}{{\mathbf{M}}^{-1}}\mathbf{w} \right)\!\!+\!\!C\sum\limits_{i=1}^{n}{{{\xi }_{i}}} \right\}, \\
 s.t. \ \ \ \ & {{y}_{i}}({{\mathbf{w}}^{T}}{{\mathbf{x}}_{i}}+b)\ge 1-{{\xi }_{i}}, \forall i,\\
 \ \ \ \ & {{\xi }_{i}}\ge 0, i=1,2,\cdots ,n, \\
 \ \ \ \ & {{({{\mathbf{x}}_{i}}-\mathbf{\bar{x}})}^{T}}\mathbf{M}({{\mathbf{x}}_{i}}-\mathbf{\bar{x}})\le 1, \forall i,\\
 \ \ \ \ & \mathbf{M}\succ 0.
\end{aligned}
\end{eqnarray}

\begin{proof}
Denote $(\mathbf{\hat{w}},\hat{b},\mathbf{\hat{\xi }},\mathbf{\hat{M}},\hat{R})$ by the optimal solution to the problem in
Eq. (\ref{eq:class_radiu-margin-based-SVM-relaxation}). Let $\mathbf{\tilde{M}}=\mathbf{\hat{M}}/{{\hat{R}}^{2}}$ and $\tilde{R}=1$. It is obvious to see that $(\mathbf{\hat{w}},\hat{b},\mathbf{\hat{\xi }},\mathbf{\tilde{M}},\tilde{R})$ is also the optimal solution to the problem in Eq. (\ref{eq:class_radiu-margin-based-SVM-relaxation}) because $F(\mathbf{\hat{w}},\hat{b},\mathbf{\hat{\xi }},\mathbf{\hat{M}},\hat{R})=F(\mathbf{\hat{w}},\hat{b},\mathbf{\hat{\xi }},\mathbf{\tilde{M}},\tilde{R})$.

Next we will show that $(\mathbf{\hat{w}},\hat{b},\mathbf{\hat{\xi }},\mathbf{\tilde{M}})$ is the optimal solution to the problem in Eq. (\ref{eq:class_radiu-margin-based-SVM-relaxation-equivalentaion}). If $(\mathbf{\hat{w}},\hat{b},\mathbf{\hat{\xi }},\mathbf{\tilde{M}})$ is not the optimal solution to Eq. (\ref{eq:class_radiu-margin-based-SVM-relaxation-equivalentaion}), there must exist some $({{\mathbf{w}}^{*}},{{b}^{*}},{{\mathbf{\xi }}^{*}},{{\mathbf{M}}^{*}})$ that satisfies all inequality constraints and $L({{\mathbf{w}}^{*}},{{b}^{*}},{{\mathbf{\xi }}^{*}},{{\mathbf{M}}^{*}})<L(\mathbf{\hat{w}},\hat{b},\mathbf{\hat{\xi }},\mathbf{\tilde{M}})$. Then we can define $\tilde{R}=1$ and have $F({{\mathbf{w}}^{*}},{{b}^{*}},{{\mathbf{\xi }}^{*}},{{\mathbf{M}}^{*}},\tilde{R})<F(\mathbf{\hat{w}},\hat{b},\mathbf{\hat{\xi }},\mathbf{\tilde{M}},\tilde{R})$, which is contradictory with the assumption that $(\mathbf{\hat{w}},\hat{b},\mathbf{\hat{\xi }},\mathbf{\tilde{M}},\tilde{R})$ is the optimal solution to Eq. (\ref{eq:class_radiu-margin-based-SVM-relaxation}). Thus, we can solve the problem in Eq. (\ref{eq:class_radiu-margin-based-SVM-relaxation-equivalentaion}) with the optimal solution $(\mathbf{\hat{w}},\hat{b},\mathbf{\hat{\xi }},\mathbf{\tilde{M}})$, and then obtain the optimal solution $(\mathbf{\hat{w}},\hat{b},\mathbf{\hat{\xi }},\mathbf{\tilde{M}},\tilde{R})$ to Eq. (\ref{eq:class_radiu-margin-based-SVM-relaxation}).
\end{proof}

Without loss of generality, we assume $\tilde{R}=1$ and seek the corresponding optimal ${\bf w}$ and ${\bf M}$  by solving Eq. (\ref{eq:class_radiu-margin-based-SVM-relaxation-equivalentaion}). Moreover, to make the model robust against outliers and noisy samples, we combine the individual inequality constraints ${{({{\mathbf{x}}_{i}}-\mathbf{\bar{x}})}^{T}}\mathbf{M}({{\mathbf{x}}_{i}}-\mathbf{\bar{x}})\le 1,i=1,2,\cdots ,n$ into one integrated inequality constraint \cite{liu2013efficient}, resulting in the following radius-margin based SVM model:
\begin{eqnarray} \label{eq:class_F-SVM-1}
\begin{aligned}
 \min \limits_{\mathbf{w},b,\mathbf{\xi },\mathbf{M}} \ \ \ \ & \frac{1}{2}\left( {{\mathbf{w}}^{T}}{{\mathbf{M}}^{-1}}\mathbf{w} \right)+C\sum\limits_{i=1}^{n}{{{\xi }_{i}}}, \\
 s.t. \ \ \ \ &  {{y}_{i}}({{\mathbf{w}}^{T}}{{\mathbf{x}}_{i}}+b)\ge 1-{{\xi }_{i}}, \forall i,\\
 \ \ \ \ & {{\xi }_{i}}\ge 0, i=1,2,\cdots ,n, \\
 \ \ \ \ & \sum\limits_{i=1}^{n}{{{({{\mathbf{x}}_{i}}-\mathbf{\bar{x}})}^{T}}\mathbf{M}({{\mathbf{x}}_{i}}-\mathbf{\bar{x}})}\le \kappa, \\
 \ \ \ \ & \mathbf{M}\succ 0.
\end{aligned}
\end{eqnarray}
By defining the scattering matrix of the training set $\mathbf{S}=\sum\nolimits_{i=1}^{n}{({{\mathbf{x}}_{i}}-\mathbf{\bar{x}}){{({{\mathbf{x}}_{i}}-\mathbf{\bar{x}})}^{T}}}$, based on the Lagrangian multiplier method \cite{boyd2004convex}, the problem expressed in Eq. (\ref{eq:class_F-SVM-1}) can be equivalently reformulated as the following F-SVM model:
\begin{eqnarray} \label{eq:class_F-SVM-2}
\begin{aligned}
 \min \limits_ {\mathbf{w},b,\mathbf{\xi },\mathbf{M}} \ \ \ \ & \frac{1}{2}\left( {{\mathbf{w}}^{T}}{{\mathbf{M}}^{-1}}\mathbf{w} \right)+C\sum\limits_{i=1}^{n}{{{\xi }_{i}}}+\rho tr\left( \mathbf{MS} \right), \\
s.t. \ \ \ \ & {{y}_{i}}({{\mathbf{w}}^{T}}{{\mathbf{x}}_{i}}+b)\ge 1-{{\xi }_{i}}, \forall i,\\
 \ \ \ \ & {{\xi }_{i}}\ge 0,i=1,2,\cdots ,n, \\
 \ \ \ \ & \mathbf{M}\succ 0.
\end{aligned}
\end{eqnarray}
where ${\rho}$ is the regularization parameter determined by ${\kappa}$. In the following, we prove that our F-SVM model is convex.

{\textbf{Theorem 3}}. The F-SVM model is a convex optimization problem.
The proof can be found in {\textbf{Appendix B}}.
\begin{figure*}
\begin{center}
\includegraphics[width=6.0in]{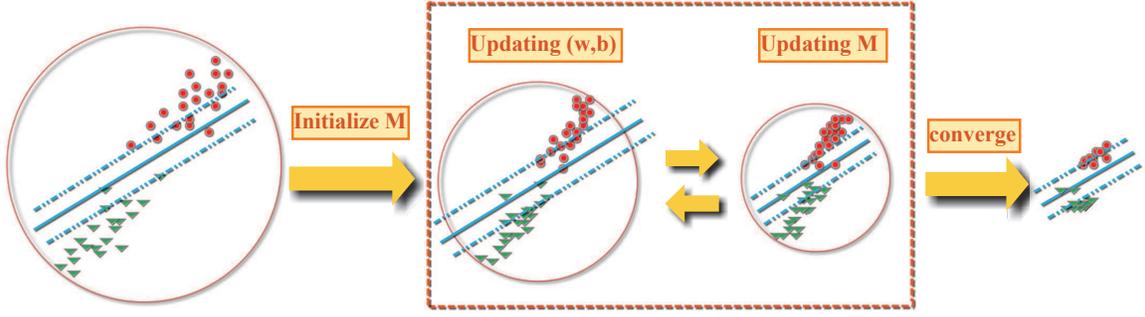}
\end{center}
 \caption{Illustration of the alternating minimization algorithm for F-SVM. First, a semi-whitened based initialization scheme is proposed to initialize ${\bf M}$. Then, the alternating minimization algorithm is adopted by updating $( {\bf w}, b)$ and ${\bf M}$ alternatively. When the algorithm converges, as shown in Fig.~\ref{fig:framework}, we can learn both a better matrix M to reduce the radius of MEB in feature space and a max-margin classifier $( {\bf w}, b)$.}
\label{fig:framework}
\end{figure*}

\subsection{Alternating minimization}
In this section, we propose an efficient alternating minimization algorithm to solve the proposed F-SVM model, as illustrated in Fig.~\ref{fig:framework}. First, a semi-whitened PCA based initialization scheme is proposed to initialize ${\bf M}$. Then, the alternating minimization algorithm is adopted by updating $( {\bf w}, b)$ and ${\bf M}$ alternatively. Fixing ${\bf M}$, the model can be reformulated as the SVM in the transformed feature space, and be solved using the off-the-shelf SVM solvers to update $( {\bf w}, b)$. Fixing $( {\bf w}, b)$, the gradient descent method \cite{boyd2004convex} is adopted to update the matrix ${\bf M}$. When the algorithm converges, as shown in Fig.~\ref{fig:framework}, we can learn both a better matrix M to reduce the radius of MEB in feature space and a max-margin classifier $( {\bf w}, b)$. The alternating optimization procedure is summarized in {\textbf{Algorithm 1}}.

\alglanguage{pseudocode}
\begin{algorithm}[htb]
\caption{The alternating minimization algorithm for F-SVM}
\label{Algm:model}
\begin{algorithmic}[1]
 \Require Training set $\left\{ \left( {{\mathbf{x}}_{i}},{{y}_{i}} \right)\left| \forall i \right. \right\}$.\
 \Ensure  Optimal ${\bf M}$ and $( {\bf w}, b)$.\
 \State $k$ = 1,
 \State Initialize \\
  \ \ \ \ ${{\mathbf{M}}_{k}}=\sqrt{{{\tau }'}}\mathbf{U\Xi }{{\mathbf{U}}^{T}}$, \\
  \ \ \ \ $\mathbf{\Xi }={diag}\{{{\left( {{\lambda }_{1}} \right)}^{-\frac{1}{2}}},{{\left( {{\lambda }_{2}} \right)}^{-\frac{1}{2}}},\cdots ,{{\left( {{\lambda }_{d
  }} \right)}^{-\frac{1}{2}}}\}$,
 \Repeat
    \State // Lines 7-9: updating $({\bf w}, b)$.
    \State Do Eigenvalue decomposition on ${{\mathbf{M}}_{k}}$: ${{\mathbf{M}}_{k}}=\mathbf{V\Sigma }{{\mathbf{V}}^{T}}$,
    \State Perform linear transformation on ${{\mathbf{x}}_{i}}$: ${{\mathbf{z}}_{i}}\!\leftarrow\!{{\mathbf{\Sigma }}^{{1}/{2}}}\!{{\mathbf{V}}^{T}}\!{{\mathbf{x}}_{i}}$,
    \State Update the SVM classifier $({\bf w}, b)$ based on ${\bf Z}$,
    \State // Lines 11-18: updating ${\bf M}$.
    \While{not converged}
        \State $\mathbf{M}={{\mathbf{M}}_{k}}$, ${\it t}=1$,
        \State Compute the gradient of ${\bf M}$: \\
        \ \ \ \ \ \ \ \ \ \ \ \ $\nabla f(\mathbf{M})=-\frac{1}{2}\mathbf{M}_{{}}^{-1}\mathbf{w}{{\mathbf{w}}^{T}}\mathbf{M}_{{}}^{-1}+\rho \mathbf{S}$,
        \State Update ${\bf M}$: $\mathbf{M}={{\mathcal{P}}_{{{\mathcal{S}}_{+}}}}\left( \mathbf{M}-t\nabla f(\mathbf{M}) \right)$,
        \State Update the stepsize ${\it t}$: $t\leftarrow \beta *t$,
    \EndWhile
    \State ${{\mathbf{M}}_{k+1}}=\mathbf{M}$,
    \State $k\leftarrow k+1$,
 \Until{${\bf M}$ and $({\bf w}, b)$ converge}
\end{algorithmic}
\end{algorithm}

\subsubsection{Initialization of M}
Because the proposed F-SVM model is convex, alternating minimization can converge to global optimum for any initialization of ${\bf M}$ and $({{\bf w}, b})$, but proper initialization is helpful in improving the computational efficiency. Thus, by further relaxing the F-SVM model in Eq. (\ref{eq:class_F-SVM-2}), we propose a semi-whitened PCA based initialization method on ${\bf M}$.

Note that ${{\mathbf{w}}^{T}}{{\mathbf{M}}^{-1}}\mathbf{w}$ is upper bounded by \cite{hom1991topics}:
\begin{eqnarray} \label{eq:class_Initial_M}
\begin{aligned}
{\bf w}^{T}{{\bf M}^{-1}}{\bf w} & =tr\left( {\bf w}{\bf w}^{T}{{\bf M}^{-1}} \right)\\
 & \le \left\| {\bf w} \right\|_{2}^{2}{{\left\|{{\bf M}^{-1}} \right\|}_{2}}\\
 & \le \left\| {\bf w} \right\|_{2}^{2}{{\left\| {{{\bf M}}^{-1}} \right\|}_{*}}.
\end{aligned}
\end{eqnarray}
where ${{\left\| \mathbf{A} \right\|}_{2}}$ and ${{\left\| \mathbf{A} \right\|}_{*}}$ denote the L$_2$-norm and the nuclear norm of a positive semi-definite matrix ${\bf A}$, respectively. Based on Eq. (\ref{eq:class_F-SVM-2}) and Eq. (\ref{eq:class_Initial_M}), by setting $\mathbf{B}={{\mathbf{M}}^{-1}}$, the subproblem on ${\bf M}$ can be rewritten as the problem on ${\bf B}$ formulated as:
\begin{eqnarray} \label{eq:class_B}
\begin{aligned}
 \min \limits_{\mathbf{B}} \ \ \ \ & L(\mathbf{B}) = {{\left\| \mathbf{B} \right\|}_{*}}+{\tau }'tr\left( {{\mathbf{B}}^{-1}}\mathbf{S} \right), \\
 s.t. \ \ \ \ & \mathbf{B}\succ 0.
\end{aligned}
\end{eqnarray}
where ${\tau }'={\rho }/{{{\left\| \mathbf{w} \right\|}^{2}}}$. The eigenvalue decomposition of ${\bf S}$ is $\mathbf{S}=\mathbf{U\Lambda }{{\mathbf{U}}^{T}}$, where $\mathbf{\Lambda }=diag({{\lambda }_{1}},{{\lambda }_{2}},\cdots ,{{\lambda }_{d}})$ (${{\lambda }_{1}}\ge {{\lambda }_{2}}\ge \cdots \ge {{\lambda }_{d}}\ge 0$), ${{\lambda }_{i}}$ and the {\it i}th column of ${\bf U}$ denote the {\it i}th eigenvalue and eigenvector, rsespectively. With ${\bf U}$ and $\mathbf{\Lambda }$, we define $\mathbf{\hat{B}}$ as:
\begin{eqnarray} \label{eq:class_15}
& \mathbf{\hat{B}}\!=\!\mathbf{U\Sigma}{{\mathbf{U}}^{T}}\!, \mathbf{\Sigma }\!=\!{diag}\{{{({\tau }'\!{{\lambda }_{1}})}^{\!1/\!2}},\cdots ,{{({\tau }'\!{{\lambda }_{d}})}^{\!1/\!2}}\}.
\end{eqnarray}

{\textbf{Theorem 4}}. Given a SPD matrix ${\bf S}$ and ${\tau }'>0$, $\mathbf{\hat{B}}$ defined in Eq. (\ref{eq:class_15}) is the optimal solution to the problem:
\begin{eqnarray} \label{eq:class_16}
\mathbf{\hat{B}}=\arg \min \limits_{\mathbf{B}} \left\{ L(\mathbf{B},{\tau }') = {{\left\| \mathbf{B} \right\|}_{*}}+{\tau }'\left( tr\left( {{\mathbf{B}}^{-1}}\mathbf{S} \right) \right) \right\}.
\end{eqnarray}
The proof can be found in {\textbf{Appendix C}}. With $\mathbf{\hat{B}}$, the initialization of ${\bf M}$ in Eq. (\ref{eq:class_F-SVM-2}) is then defined as:
\begin{eqnarray} \label{eq:class_17}
\begin{aligned}
 {{\bf M}_{0}} \! = \! \sqrt{{\tau }'}\mathbf{U\Xi }{{\mathbf{U}}^{T}}, \mathbf{\Xi } \!=\! {diag}\{{\left( {{\lambda }_{1}} \right)}^{\!-1\!/\!2},\cdots ,{{\left( {{\lambda }_{d}} \right)}^{\!-\!1/\!2}}\}.
\end{aligned}
\end{eqnarray}
Noted that we assume that ${{\left\| \mathbf{w} \right\|}^{2}}$ is known for the initialization of ${\bf M}$. From Eq. (\ref{eq:class_17}), ${{\left\| \mathbf{w} \right\|}^{2}}$ only affects the scale factor $\sqrt{{{\tau }'}}$ to the linear transformation. Thus, we simply let ${{\left\| \mathbf{w} \right\|}^{2}}=1$ in our implementation.

It is interesting to point out that ${{\bf M}_{0}}$ in Eq. (\ref{eq:class_17}) implies a semi-whitening PCA transformation because ${{\mathbf{M}}_{0}}=\mathbf{U}{{\mathbf{\Lambda }}^{-1/2}}{{\mathbf{U}}^{T}}$, where the linear transformation can then be defined as ${{\mathbf{\Lambda }}^{-1/4}}{{\mathbf{U}}^{T}}{{\mathbf{x}}_{i}}$. In literature \cite{hyvarinen2000independent}, ${{\mathbf{\Lambda }}^{-1/2}}{{\mathbf{U}}^{T}}{{\mathbf{x}}_{i}}$ was called the PCA whitening transformation, and data whitening has been widely exploited in many applications, e.g., face recognition, object detection, and image classification \cite{yang2005ica, krizhevsky2009learning, hariharan2012discriminative, girshick2013training}. Considering its connection with data whitening, semi-whitening is also expected to be effective in other tasks and applications.

\subsubsection{The subproblem on $( {\bf w}, b)$}

Given ${\bf M}$, the F-SVM model can be formulated as:
\begin{eqnarray} \label{eq:class_18}
\begin{aligned}
 \min \limits_{\mathbf{w},b,\mathbf{\xi }} \ \ \ \ & \frac{1}{2}{{\mathbf{w}}^{T}}\mathbf{Bw}+C\sum\limits_{i=1}^{n}{{{\xi }_{i}}}, \\
 s.t. \ \ \ \ & {{y}_{i}}({{\mathbf{w}}^{T}}{{\mathbf{x}}_{i}}+b)\ge 1-{{\xi }_{i}}, \forall i,\\
 \ \ \ \ & {{\xi }_{i}}\ge 0,i=1,2,\cdots ,n.
\end{aligned}
\end{eqnarray}
where $\mathbf{B}={{\mathbf{M}}^{-1}}$. The eigenvalue decomposition of ${\bf B}$ is $\mathbf{B}=\mathbf{V\Sigma }{{\mathbf{V}}^{T}}$. By introducing $\mathbf{B}={{\mathbf{L}}^{T}}\mathbf{L}$, the transformation matrix ${\bf L}$ can be rewritten as to $\mathbf{L}={{\mathbf{\Sigma }}^{\frac{1}{2}}}{{\mathbf{V}}^{T}}$. Let ${{\mathbf{z}}_{i}}={{\mathbf{\Sigma }}^{-\frac{1}{2}}}{{\mathbf{V}}^{T}}{{\mathbf{x}}_{i}}$ and $\mathbf{v}=\mathbf{Lw}$. With simple algebra, the problem in Eq. (\ref{eq:class_18}) can be reformulated as:
\begin{eqnarray} \label{eq:class_19}
\begin{aligned}
 \min \limits_{\mathbf{v},b,\mathbf{\xi }} \ \ \ \ & \frac{1}{2}{{\mathbf{v}}^{T}}\mathbf{v}+C\sum\limits_{i=1}^{n}{{{\xi }_{i}}}, \\
 s.t. \ \ \ \ & {{y}_{i}}({{\mathbf{v}}^{T}}{{\mathbf{z}}_{i}}+b)\ge 1-{{\xi }_{i}}, \forall i,\\
 \ \ \ \ & {{\xi }_{i}}\ge 0,i=1,2,\cdots ,n.
\end{aligned}
\end{eqnarray}
which can be solved using the off-the-shelf SVM solvers. Given the solution ${\bf v}$, $\mathbf{w}=\mathbf{V}{{\mathbf{\Sigma }}^{-\frac{1}{2}}}\mathbf{v}$ can then be obtained.

\subsubsection{The subproblem on M}

Given $( {\bf w}, b)$, the sub-problem on ${\bf M}$ can be reformulated as:
\begin{eqnarray} \label{eq:class_20}
\begin{aligned}
 \min \limits_{\mathbf{M}} \ \ \ \ & f(\mathbf{M})=\frac{1}{2}\left( {{\mathbf{w}}^{T}}{{\mathbf{M}}^{-1}}\mathbf{w} \right)+\rho tr\left( \mathbf{MS} \right), \\
 s.t. \ \ \ \ & \mathbf{M}\succ 0.
\end{aligned}
\end{eqnarray}
Since the objective function in Eq. (\ref{eq:class_20}) is convex and differentiable with respect to ${\bf M}$, the gradient projection method \cite{boyd2004convex} is adopted to update ${\bf M}$. According to \cite{petersen2008matrix}, the gradient of $f({\bf M})$ can be obtained by:
\begin{eqnarray}\label{eq:class_21}
\nabla f(\mathbf{M})=-\frac{1}{2}{{\mathbf{M}}^{-1}}\mathbf{w}{{\mathbf{w}}^{T}}{{\mathbf{M}}^{-1}}+\rho \mathbf{S}.
\end{eqnarray}
As presented in {\textbf {Algorithm 1}}, we use gradient projection
\begin{eqnarray}\label{eq:class_22}
\mathbf{M}={{\mathcal{P}}_{{{\mathcal{S}}_{+}}}}\left( \mathbf{M}-t\nabla f(\mathbf{M}) \right)
\end{eqnarray}
to update ${\bf M}$ by choosing proper stepsize ${\it t}$ and gradually decreasing it along with iterations, where ${{\mathcal{P}}_{{{\mathcal{S}}_{+}}}}\left( \cdot  \right)$ projects a matrix onto the cone of positive semidefinite matrices.

\subsection{Discussion}
The proposed F-SVM method has several interesting advantages while compared with the other radius-margin based SVMs, e.g., RMM \cite{shivaswamy2010maximum}, MR-SVM \cite{do2009feature}, R-SVM$^{+}$ \cite{do2013convex}, R-SVM$_{\mu}^{+}$ \cite{do2013convex} and M-SVM \cite{zhu2012learning}. RMM \cite{shivaswamy2010maximum} is suggested to maximize the margin while restricting the spread of the data along the direction perpendicular to the separating hyperplane, while our F-SVM is proposed to minimize the convex relaxation of the radius-margin ratio. The generalization error is bounded by the radius and margin ratio, and the radius is determined by the spread along all possible directions rather than only the direction perpendicular to the separating hyperplane, making F-SVM theoretically more promising.

MR-SVM \cite{do2009feature}, R-SVM$^{+}$ \cite{do2013convex} and R-SVM$_{\mu}^{+}$ \cite{do2013convex} aim to learn the diagonal feature transformation ${{\mathbf{D}}_{\mathbf{\mu }}}=Diag\left( \mathbf{\mu } \right)$ with ${{\mu }_{k}}\ge 0$, while F-SVM is developed for joint learning of feature transformation and SVM classifier. Both R-SVM$^{+}$ and R-SVM$_{\mu}^{+}$ need to solve a Quadratically Constrained Quadratic Programming (QCQP) optimization problem, which is computationally expensive than the alternating minimization method used in our F-SVM. Moreover, R-SVM$^{+}$ and R-SVM$_{\mu}^{+}$ adopted a tighter approximation ${{R}_{O}}$ of the radius. In F-SVM, a new approximation $\bar{R}$ of the radius is proposed, which is also tighter than that used in MR-SVM \cite{do2009feature}. Moreover, the individual inequality constraints on $\bar{R}$ are combined to improve the robustness against outliers. It is interesting to note that we have:
\begin{eqnarray} \label{eq:class_23}
\sum\limits_{i,j}{{{({{\mathbf{x}}_{i}}-{{\mathbf{x}}_{j}})}^{T}}\!\mathbf{M}\!({{\mathbf{x}}_{i}}-{{\mathbf{x}}_{j}})}\!=\!tr\left( \mathbf{M}{{\mathbf{S}}_{t}} \right)\!=\!4n\left( tr\left( \mathbf{MS} \right) \right),
\end{eqnarray}
where ${{\mathbf{S}}_{t}}=\sum\nolimits_{i,j}{({{\mathbf{x}}_{i}}-{{\mathbf{x}}_{j}}){{({{\mathbf{x}}_{i}}-{{\mathbf{x}}_{j}})}^{T}}}$. Eq. (\ref{eq:class_23}) indicates that, if all the inequality constraints on ${{R}_{O}}$, i.e., ${{\left\| {{D}_{\sqrt{\mathbf{\mu }}}}{{\mathbf{x}}_{i}}-{{D}_{\sqrt{\mathbf{\mu }}}}{{\mathbf{x}}_{j}} \right\|}^{2}}\le r$, are combined into one integrated inequality constraint:
\begin{eqnarray} \label{eq:class_24}
\sum\limits_{i,j}\!\!{{{({{\mathbf{x}}_{i}}-{{\mathbf{x}}_{j}})}^{T}}\!D_{\sqrt{\mathbf{\mu }}}^{T}{{D}_{\sqrt{\mathbf{\mu }}}}\!({{\mathbf{x}}_{i}}-{{\mathbf{x}}_{j}})}\!=\!tr\left( D_{\sqrt{\mathbf{\mu }}}^{T}{{D}_{\sqrt{\mathbf{\mu }}}}{{\mathbf{S}}_{t}} \right)\!\le\!{\kappa }'.
\end{eqnarray}
Let ${\kappa }'=4n\kappa$ and $\mathbf{M}=D_{\sqrt{\mathbf{\mu }}}^{T}{{D}_{\sqrt{\mathbf{\mu }}}}$. One can see that the integrated inequality constraint will be equivalent with that adopted in Eq. (\ref{eq:class_F-SVM-1}).

MSVM \cite{zhu2012learning} was developed for simultaneous learning of the linear transformation and SVM classifier, but the MSVM model is non-convex and solved using gradient projection. Moreover, although Zhu et al. \cite{zhu2012learning} claimed that $R=\bar{R}$, as discussed in Section 3.1, $\bar{R}$ is only a lower bound of $R$  and counter examples can be easily found to illustrate $R\ne \bar{R}$. Compared with MSVM \cite{zhu2012learning}, the F-SVM model is convex and robust against noise and outliers, and can be efficient solved using the optimization method introduced in Section 3.2.

\section{Kernelization of F-SVM}
With the incorporation of kernel principal component analysis, linear F-SVM can be extended to kernel version for nonlinear classification. First, we show that kernel SVM is equivalent to perform linear SVM in the kernel PCA space. Then, kernel F-SVM is introduced by conducting linear F-SVM in the kernel PCA space.

Let the kernel function be $K({{\mathbf{x}}_{i}},{{\mathbf{x}}_{j}})=\varphi {{({{\mathbf{x}}_{i}})}^{T}}\varphi ({{\mathbf{x}}_{j}})$, where $\varphi \left( \mathbf{x} \right)$
defines an implicit mapping from the data space to high or infinite dimensional feature space. For the training set $\mathcal{S}=\left\{ \left( {{\mathbf{x}}_{1}},{{y}_{1}} \right),...,\left( {{\mathbf{x}}_{n}},{{y}_{n}} \right) \right\}$, we use $\mathbf{W}=\left[ {{\mathbf{w}}_{1}},{{\mathbf{w}}_{2}},...,{{\mathbf{w}}_{D}} \right]$ to denote all the PCA eigenvectors corresponding to positive eigenvalues. Let $\mathbf{\bar{W}}$ be a set of basis vectors in the complementary space of $\mathbf{W}$. Assuming the training set is centered, for any ${{\bf x}_{i}}$, we have ${{\mathbf{\bar{W}}}^{T}}\varphi ({{\mathbf{x}}_{i}})=0$, and thus can get:
\begin{eqnarray} \label{eq:class_25}
\begin{aligned}
 K({{\mathbf{x}}_{i}},{{\mathbf{x}}_{j}})&\!=\!\varphi {{({{\mathbf{x}}_{i}})}^{T}}\mathbf{W}{{\mathbf{W}}^{T}}\varphi ({{\mathbf{x}}_{j}})\!+\!\varphi {{({{\mathbf{x}}_{i}})}^{T}}\mathbf{\bar{W}}{{\mathbf{\bar{W}}}^{T}}\varphi ({{\mathbf{x}}_{j}})\\
&=\varphi {{({{\mathbf{x}}_{i}})}^{T}}\mathbf{W}{{\mathbf{W}}^{T}}\varphi ({{\mathbf{x}}_{j}}).
\end{aligned}
\end{eqnarray}
Let ${{\mathbf{f}}_{i}}={{\mathbf{W}}^{T}}\varphi ({{\mathbf{x}}_{i}})$. The dual problem of SVM in the kernel PCA space can be formulated as:
\begin{eqnarray} \label{eq:class_26}
\begin{aligned}
  \max \limits_{\mathbf{\alpha }} \ \ \ \ & Q\left( \mathbf{\alpha } \right)=\sum\nolimits_{i=1}^{n}{{{\alpha}_{i}}}-\frac{1}{2}\sum\nolimits_{i,j=1}^{n}{{{\alpha}_{i}}{{\alpha}_{j}}{{y}_{i}}{{y}_{j}}\left\langle {{\mathbf{f}}_{i}},{{\mathbf{f}}_{j}} \right\rangle }, \\
  s.t. \ \ \ \ & \sum\nolimits_{i=1}^{n}{{{\alpha }_{i}}{{y}_{i}}} = 0, \\
  \ \ \ \ & 0\le {{\alpha }_{i}}\le C,\ \ i=1,...,n.
\end{aligned}
\end{eqnarray}
where $\left\langle {{\mathbf{f}}_{i}},{{\mathbf{f}}_{j}} \right\rangle =K\left( {{\mathbf{x}}_{i}},{{\mathbf{x}}_{j}} \right)$. Therefore, kernel SVM is equivalent to performing linear SVM in the kernel PCA space. To extend F-SVM to its kernelized version, we first project each training sample ${{\bf x}_i}$ to the kernel PCA space ${{\mathbf{f}}_{i}}={{\mathbf{W}}^{T}}\varphi ({{\mathbf{x}}_{i}})$, and then solve the following F-SVM model:
\begin{eqnarray} \label{eq:class_27}
\begin{aligned}
 \min \limits_{\mathbf{w},b,\mathbf{\xi },\mathbf{M}}  \ \ \ \ &  \frac{1}{2}\left( {{\mathbf{w}}^{T}}{{\mathbf{M}}^{-1}}\mathbf{w} \right)+C\sum\limits_{i=1}^{n}{{{\xi}_{i}}}+\rho tr\left( \mathbf{M}{{\mathbf{S}}_{f}} \right), \\
 s.t.  \ \ \ \ & {{y}_{i}}({{\mathbf{w}}^{T}}{{\mathbf{f}}_{i}}+b)\ge 1-{{\xi }_{i}}, \forall i,\\
 \ \ \ \ & {{\xi }_{i}}\ge 0,i=1,2,\cdots ,n, \\
 \ \ \ \ & \mathbf{M}\succ 0.
\end{aligned}
\end{eqnarray}
where ${{\mathbf{S}}_{f}}=\sum\nolimits_{i=1}^{n}{{{\mathbf{f}}_{i}}\mathbf{f}_{i}^{T}}$. {\textbf {Algorithm 1}} can be adopted to solve the model in Eq. (27). In our implementation, instead of using all the eigenvectors, we also consider to employ the PCA eigenvectors corresponding to first {\it d} largest eigenvalues, i.e., $\mathbf{W}=\left[ {{\mathbf{w}}_{1}},{{\mathbf{w}}_{2}},...,{{\mathbf{w}}_{D}} \right]$, and Section 5.2 reports the empirical result on the influence of {\it d} on classification accuracy.

\section{Experiments}
In this section, we use both the UCI machine learning datasets and the Labeled Faces in the Wild (LFW) database to evaluate the proposed F-SVM method, and compare our F-SVM with the competing methods, including SVM and several representative radius-margin based SVM methods, i.e., RMM \cite{shivaswamy2010maximum}, R-SVM$^{+}$ \cite{do2013convex} and R-SVM$_{\mu}^{+}$ \cite{do2013convex}. MR-SVM \cite{do2009feature} and MSVM \cite{zhu2012learning} are not considered in our experiments because their source codes are not publicly available.
The 10-fold cross validation (CV) is adopted to determine the optimal values of hyper-parameters for each method. The mean classification accuracy is adopted by averaging the 100 runs of the 10-fold CV. The methods are evaluated by two performance indicators: accuracy and training time (seconds, {\it s}).

\begin{table}[!t] \scriptsize 
\renewcommand{\arraystretch}{1}
\caption{Summary of the UCI datasets used in the experiments.}
\label{table1} \centering
\begin{tabular}{|c|c|c|c|}
\hline
 Dataset & \# of samples  & \# of classes & \# of attributes \\
\hline
Breast cancer           & 286   & 2  & 9 \\
\hline
Diabetes   & 768   & 2  & 8 \\
\hline
Solar Flare             & 144   & 3  & 9 \\
\hline
German                  & 1000  & 2  & 20 \\
\hline
Heart                   & 270   & 2  & 13 \\
\hline
Image                   & 2310  & 7  & 19 \\
\hline
Ringnorm                & 7400  & 2  & 20 \\
\hline
Splice                  & 3190  & 3  & 60 \\
\hline
Thyroid                 & 215   & 3  & 5 \\
\hline
Twonorm                 & 7400  & 2  & 20 \\
\hline
Waveform                & 5000  & 3  & 21 \\
\hline
\end{tabular}
\end{table}

\subsection{Evaluation on linear F-SVM using the UCI datasets}
We evaluate the performance of linear F-SVM on the 11 datasets from the UCI machine learning repository, where the reason to choose them is that they had been widely adopted for evaluating SVM and kernel methods \cite{ratsch2001soft, scholkopft1999fisher, cawley2003efficient}. Table \ref{table1} provides a brief summary of these UCI datasets, which includes 6 2-class problems and 5 multi-class problems. Tables \ref{table2} and \ref{table3} list the mean classification accuracy and training time of five linear classifiers, i.e., linear SVM, linear RMM \cite{shivaswamy2010maximum}, linear R-SVM$^{+}$ \cite{do2013convex}, linear R-SVM$_{\mu}^{+}$ \cite{do2013convex}, and linear F-SVM.
RMM \cite{shivaswamy2010maximum}, R-SVM$^{+}$ \cite{do2013convex}, R-SVM$_{\mu}^{+}$ \cite{do2013convex}, and F-SVM consider both margin and radius information, while SVM only considers margin. As shown in Table \ref{table2}, the radius-margin based SVM methods generally outperform SVM in terms of classification accuracy, which indicates that the incorporation of radius can improve the classification performance. As listed in Table \ref{table3}, the training time of SVM is much less than the other four methods, indicating that the introduction of radius makes the model more complex to train.

\begin{table}[!t] \scriptsize 
\renewcommand{\arraystretch}{1}
\caption{Comparison of the average classification accuracy (\%) of linear SVM, linear RMM \cite{shivaswamy2010maximum}, linear R-SVM$^{+}$ \cite{do2013convex}, linear R-SVM$_{\mu}^{+}$ \cite{do2013convex}, and linear F-SVM.}
\label{table2} \centering
\begin{tabular}{ |c|c|c|c|c|c| }
\hline
 Dataset    & SVM    & RMM    & R-SVM$^{+}$    & R-SVM$_{\mu}^{+}$    & F-SVM \\
\hline
Breast cancer           & 71.40   & 70.19  & 71.54  & 71.14  & \textbf{71.68} \\
\hline
Diabetes   & 76.57   & 76.29  & 76.67  &76.42   &\textbf{77.00}  \\
\hline
Solar Flare             & 67.66   & 67.38  & 67.66  &67.66   &\textbf{67.69}  \\
\hline
German                  & 75.58   & 75.99  & 76.01  &75.87   &\textbf{76.04}  \\
\hline
Heart                   & 83.61   & 83.64  & 83.83  &83.96   &\textbf{84.02}  \\
\hline
Image                   & 83.77   & 84.12  & \textbf{84.39}  &83.97   &84.32  \\
\hline
Ringnorm                & 75.41   & 75.78  & 75.63  &75.43   &\textbf{77.05}  \\
\hline
Splice                  & 84.54   & \textbf{85.05}  & 84.67  &84.74   &84.81  \\
\hline
Thyroid                 & 89.76   & 91.19  & \textbf{91.23}  &90.09   &86.81  \\
\hline
Twonorm                 & 96.92   & \textbf{97.79}  & 97.41  &97.39   &97.08  \\
\hline
Waveform                & 86.95   & \textbf{88.54}  & 88.51  &86.88   &86.76  \\
\hline
\end{tabular}
\end{table}

\begin{table}[!t] \scriptsize  
\renewcommand{\arraystretch}{1}
\caption{Comparison of the training time (${\it s}$) of linear SVM, linear RMM \cite{shivaswamy2010maximum}, linear R-SVM$^{+}$ \cite{do2013convex}, linear R-SVM$_{\mu}^{+}$ \cite{do2013convex}, and linear F-SVM.}
\label{table3} \centering
\begin{tabular}{ |c|c|c|c|c|c| }
\hline
 Dataset    & SVM    & RMM    & R-SVM$^{+}$    & R-SVM$_{\mu}^{+}$    & F-SVM \\
\hline
Breast cancer           & $\!\!6.20\!\!\times\!\!10^{\!-\!3}$   & $\!\!3.70\!\!\times\!\!10^{\!+\!1}$  & $\!\!2.29\!\!\times\!\!10^{\!+\!1}$  & $\!\!0.20\!\!\times\!\!10^{\!+\!0}$  & $\!\!6.60\!\!\times\!\!10^{\!-\!3}$ \\
\hline
Diabetes                & $\!\!1.12\!\!\times\!\!10^{\!-\!2}$   & $\!\!2.63\!\!\times\!\!10^{\!+\!2}$  & $\!\!2.80\!\!\times\!\!10^{\!+\!2}$  & $\!\!1.20\!\!\times\!\!10^{\!+\!3}$  & $\!\!1.38\!\!\times\!\!10^{\!-\!2}$ \\
\hline
Solar Flare             & $\!\!7.79\!\!\times\!\!10^{\!-\!4}$   & $\!\!1.38\!\!\times\!\!10^{\!+\!1}$  & $\!\!3.41\!\!\times\!\!10^{\!+\!1}$  & $\!\!4.61\!\!\times\!\!10^{\!+\!8}$  & $\!\!6.60\!\!\times\!\!10^{\!-\!4}$ \\
\hline
German                  & $\!\!4.83\!\!\times\!\!10^{\!-\!2}$   & $\!\!3.77\!\!\times\!\!10^{\!+\!2}$  & $\!\!7.38\!\!\times\!\!10^{\!+\!2}$  & $\!\!1.55\!\!\times\!\!10^{\!+\!2}$  & $\!\!5.95\!\!\times\!\!10^{\!-\!2}$ \\
\hline
Heart                   & $\!\!1.50\!\!\times\!\!10^{\!-\!3}$   & $\!\!3.06\!\!\times\!\!10^{\!+\!1}$  & $\!\!5.09\!\!\times\!\!10^{\!+\!1}$  & $\!\!0.29\!\!\times\!\!10^{\!+\!0}$  & $\!\!1.80\!\!\times\!\!10^{\!-\!3}$ \\
\hline
Image                   & $\!\!2.29\!\!\times\!\!10^{\!+\!0}$   & $\!\!2.21\!\!\times\!\!10^{\!+\!3}$  & $\!\!5.79\!\!\times\!\!10^{\!+\!3}$  & $\!\!1.48\!\!\times\!\!10^{\!+\!3}$  & $\!\!2.40\!\!\times\!\!10^{\!+\!0}$ \\
\hline
Ringnorm                & $\!\!1.42\!\!\times\!\!10^{\!+\!1}$   & $\!\!7.46\!\!\times\!\!10^{\!+\!3}$  & $\!\!2.63\!\!\times\!\!10^{\!+\!4}$  & $\!\!1.13\!\!\times\!\!10^{\!+\!3}$  & $\!\!1.42\!\!\times\!\!10^{\!+\!1}$ \\
\hline
Splice                  & $\!\!4.02\!\!\times\!\!10^{\!+\!1}$   & $\!\!2.77\!\!\times\!\!10^{\!+\!3}$  & $\!\!1.09\!\!\times\!\!10^{\!+\!4}$  & $\!\!5.69\!\!\times\!\!10^{\!+\!3}$  & $\!\!1.12\!\!\times\!\!10^{\!+\!0}$ \\
\hline
Thyroid                 & $\!\!8.70\!\!\times\!\!10^{\!-\!3}$   & $\!\!2.06\!\!\times\!\!10^{\!+\!1}$  & $\!\!2.79\!\!\times\!\!10^{\!+\!1}$  & $\!\!0.21\!\!\times\!\!10^{\!+\!0}$  & $\!\!1.59\!\!\times\!\!10^{\!-\!1}$ \\
\hline
Twonorm                 & $\!\!5.99\!\!\times\!\!10^{\!+\!0}$   & $\!\!3.14\!\!\times\!\!10^{\!+\!1}$  & $\!\!9.91\!\!\times\!\!10^{\!+\!3}$  & $\!\!1.16\!\!\times\!\!10^{\!+\!2}$  & $\!\!3.16\!\!\times\!\!10^{\!-\!1}$ \\
\hline
Waveform                & $\!\!1.30\!\!\times\!\!10^{\!-\!1}$   & $\!\!6.54\!\!\times\!\!10^{\!+\!4}$  & $\!\!5.20\!\!\times\!\!10^{\!+\!3}$  & $\!\!1.19\!\!\times\!\!10^{\!+\!1}$  & $\!\!6.71\!\!\times\!\!10^{\!-\!1}$ \\
\hline
\end{tabular}
\end{table}

We further compare linear F-SVM with the competing methods. From Table \ref{table2}, F-SVM achieves higher classification accuracy than SVM on 9, RMM \cite{shivaswamy2010maximum} on 7, R-SVM$^{+}$ \cite{do2013convex} on 7, and R-SVM$_{\mu}^{+}$ \cite{do2013convex} on 8 of the 11 datasets. The better classification accuracy of our F-SVM should be attributed to that: $(i)$ compared with SVM, F-SVM incorporates radius in the convex model; $(ii)$ unlike RMM \cite{shivaswamy2010maximum}, our F-SVM considers the spread along all directions rather than only the direction perpendicular to the separating hyperplane. $(iii)$ instead of feature reweighting and selection in R-SVM$^{+}$ \cite{do2013convex} and R-SVM$_{\mu}^{+}$ \cite{do2013convex}, general linear transformation is learned in F-SVM.
To improve the efficiency of F-SVM in training, we adopt the warm-start strategy, where the solution $( {\bf w}, b)$ of the previous iteration is used as the initialization of the next iteration. From Table \ref{table3}, one can see that our F-SVM is only a little slower than SVM, but is much more efficient than the other competing methods in training. F-SVM is about $10^3\!\!\sim\!\!10^4$ times faster than RMM \cite{shivaswamy2010maximum}, R-SVM$^{+}$ and R-SVM$_{\mu}^{+}$ in \cite{do2013convex}. In summary, F-SVM obtains the best classification accuracy among all competing methods, and is more efficient in training than the other radius-margin based SVM methods.

\begin{table}[!h]  \scriptsize 
\renewcommand{\arraystretch}{1}
\caption{Comparison of the average classification accuracy (\%) of kernel SVM, kernel RMM \cite{shivaswamy2010maximum}, kernel R-SVM$^{+}$ \cite{do2013convex}, kernel R-SVM$_{\mu}^{+}$ \cite{do2013convex}, and kernel F-SVM.}
\label{table4} \centering
\begin{tabular}{ | c | c | c | c | c | c | }
\hline
 Dataset    & SVM    & RMM    & R-SVM$^{+}$    & R-SVM$_{\mu}^{+}$    & F-SVM \\
\hline
Breast cancer           & 73.74   &\textbf{74.15}  & 71.54  & 71.14  & 73.95 \\
\hline
Diabetes                & 76.83   & 74.97  & 77.05  & 76.80  &\textbf{78.84}  \\
\hline
Solar Flare             & 67.64   & 66.33  &\textbf{67.66}  & 67.54  &\textbf{67.66} \\
\hline
German                  & 76.36   & 76.58  & 76.01  & 75.93  &\textbf{76.90}  \\
\hline
Heart                   & 83.43   & 82.19  & 83.83  &83.96   &\textbf{84.25}  \\
\hline
Image                   & 97.14   & 96.41  & 84.39  &83.97   &\textbf{96.93}  \\
\hline
Ringnorm                & 98.41   & 86.16  & 75.63  &75.43   &\textbf{98.58}  \\
\hline
Splice                  & 90.16   & 89.34  & 84.85  &84.97   &\textbf{90.55} \\
\hline
Thyroid                 & 95.91   & 95.86  & 91.23  &91.31   &\textbf{96.13}  \\
\hline
Twonorm                 & 97.59   & 95.43  & 97.41  &97.39   &\textbf{97.79}  \\
\hline
Waveform                & 89.75   &\textbf{92.60}  & 88.51  &89.38   &90.95  \\
\hline
\end{tabular}
\end{table}

\begin{table}[!t]  \scriptsize 
\renewcommand{\arraystretch}{1}
\caption{Comparison of the training time (${\it s}$) of kernel SVM, kernel RMM \cite{shivaswamy2010maximum}, kernel R-SVM$^{+}$ \cite{do2013convex}, kernel R-SVM$_{\mu}^{+}$ \cite{do2013convex}, and kernel F-SVM.}
\label{table5} \centering
\begin{tabular}{ |c|c|c|c|c|c| }
\hline
 Dataset    & SVM    & RMM    & R-SVM$^{+}$    & R-SVM$_{\mu}^{+}$    & F-SVM \\
\hline
Breast cancer           & $\!\!1.90\!\!\times\!\!10^{\!-\!3}$   & $\!\!1.00\!\!\times\!\!10^{\!+\!3}$  & $\!\!2.24\!\!\times\!\!10^{\!+\!1}$  & $\!\!1.83\!\!\times\!\!10^{\!+\!1}$  & $\!\!6.60\!\!\times\!\!10^{\!-\!3}$ \\
\hline
Diabetes   & $\!\!1.11\!\!\times\!\!10^{\!-\!2}$   & $\!\!5.70\!\!\times\!\!10^{\!+\!2}$  & $\!\!3.47\!\!\times\!\!10^{\!+\!1}$  & $\!\!5.02\!\!\times\!\!10^{\!+\!1}$  & $\!\!2.57\!\!\times\!\!10^{\!-\!2}$ \\
\hline
Solar Flare             & $\!\!1.10\!\!\times\!\!10^{\!-\!3}$   & $\!\!9.01\!\!\times\!\!10^{\!+\!1}$  & $\!\!3.52\!\!\times\!\!10^{\!+\!1}$  & $\!\!4.58\!\!\times\!\!10^{\!+\!3}$  & $\!\!1.80\!\!\times\!\!10^{\!-\!3}$ \\
\hline
German                  & $\!\!4.70\!\!\times\!\!10^{\!-\!2}$   & $\!\!1.17\!\!\times\!\!10^{\!+\!3}$  & $\!\!5.30\!\!\times\!\!10^{\!+\!2}$  & $\!\!1.83\!\!\times\!\!10^{\!+\!3}$  & $\!\!6.08\!\!\times\!\!10^{\!-\!2}$ \\
\hline
Heart                   & $\!\!1.00\!\!\times\!\!10^{\!-\!3}$   & $\!\!3.46\!\!\times\!\!10^{\!+\!1}$  & $\!\!5.08\!\!\times\!\!10^{\!+1\!}$  & $\!\!1.15\!\!\times\!\!10^{\!+\!3}$  & $\!\!1.20\!\!\times\!\!10^{\!-\!3}$ \\
\hline
Image                   & $\!\!5.30\!\!\times\!\!10^{\!-\!2}$   & $\!\!4.31\!\!\times\!\!10^{\!+\!3}$  & $\!\!3.44\!\!\times\!\!10^{\!+\!3}$  & $\!\!1.59\!\!\times\!\!10^{\!+\!3}$  & $\!\!5.98\!\!\times\!\!10^{\!-\!2}$ \\
\hline
Ringnorm                & $\!\!3.08\!\!\times\!\!10^{\!-\!1}$   & $\!\!2.93\!\!\times\!\!10^{\!+\!4}$  & $\!\!3.59\!\!\times\!\!10^{\!+\!4}$  & $\!\!6.25\!\!\times\!\!10^{\!+\!1}$  & $\!\!3.75\!\!\times\!\!10^{\!-\!1}$ \\
\hline
Splice                  & $\!\!6.28\!\!\times\!\!10^{\!-\!1}$   & $\!\!6.49\!\!\times\!\!10^{\!+\!3}$  & $\!\!3.13\!\!\times\!\!10^{\!+\!4}$  & $\!\!6.69\!\!\times\!\!10^{\!+\!3}$  & $\!\!6.07\!\!\times\!\!10^{\!-\!1}$ \\
\hline
Thyroid                 & $\!\!4.22\!\!\times\!\!10^{\!-\!4}$   & $\!\!4.51\!\!\times\!\!10^{\!+\!1}$  & $\!\!4.13\!\!\times\!\!10^{\!+\!1}$  & $\!\!7.13\!\!\times\!\!10^{\!+\!0}$  & $\!\!5.95\!\!\times\!\!10^{\!-\!4}$ \\
\hline
Twonorm                 & $\!\!2.81\!\!\times\!\!10^{\!-\!1}$   & $\!\!3.26\!\!\times\!\!10^{\!+\!4}$  & $\!\!1.97\!\!\times\!\!10^{\!+\!4}$  & $\!\!2.62\!\!\times\!\!10^{\!+\!1}$  & $\!\!3.83\!\!\times\!\!10^{\!-\!1}$ \\
\hline
Waveform                & $\!\!5.79\!\!\times\!\!10^{\!-\!1}$   & $\!\!4.24\!\!\times\!\!10^{\!+\!4}$  & $\!\!1.04\!\!\times\!\!10^{\!+\!4}$  & $\!\!3.18\!\!\times\!\!10^{\!+\!1}$  & $\!\!6.19\!\!\times\!\!10^{\!-\!1}$ \\
\hline
\end{tabular}
\end{table}

\subsection{Evaluation on kernel F-SVM using the UCI datasets}

\begin{figure}
  \centering
   \subfigure[Breast cancer]{
     \label{fig:subfig:a}
     \includegraphics[width=1.67in]{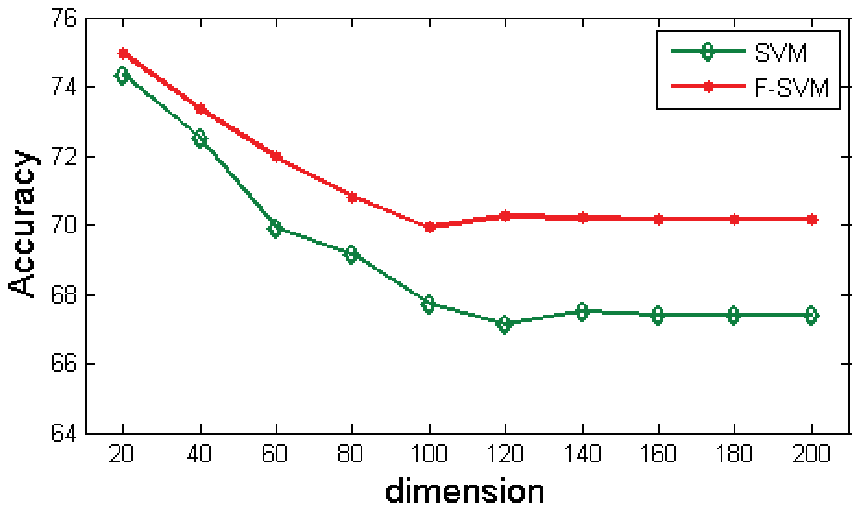}}
   \subfigure[Thyoid]{
     \label{fig:subfig:b} 
     \includegraphics[width=1.67in]{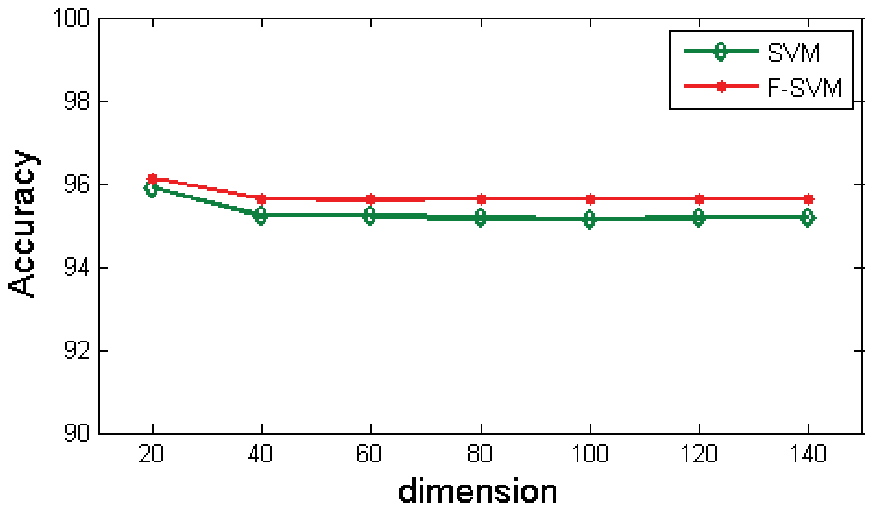}}


   \subfigure[Heart]{
     \label{fig:subfig:c} 
     \includegraphics[width=1.67in]{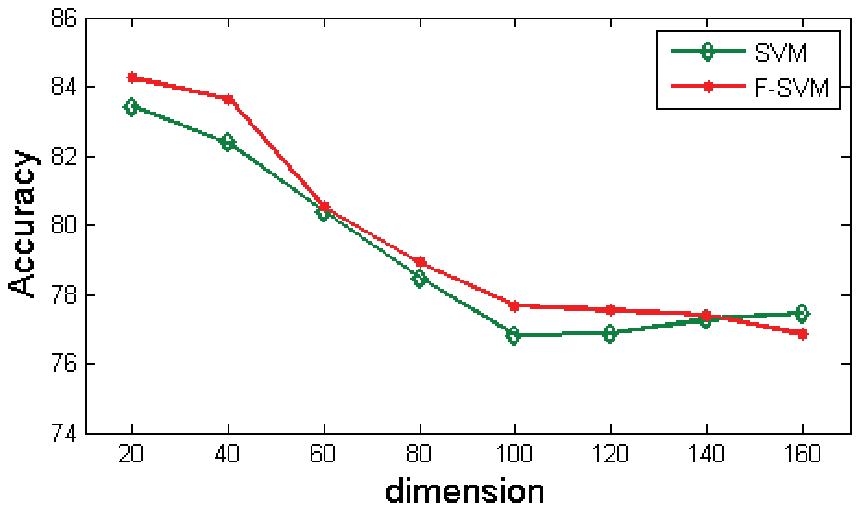}}
   \subfigure[German]{
     \label{fig:subfig:d} 
     \includegraphics[width=1.67in]{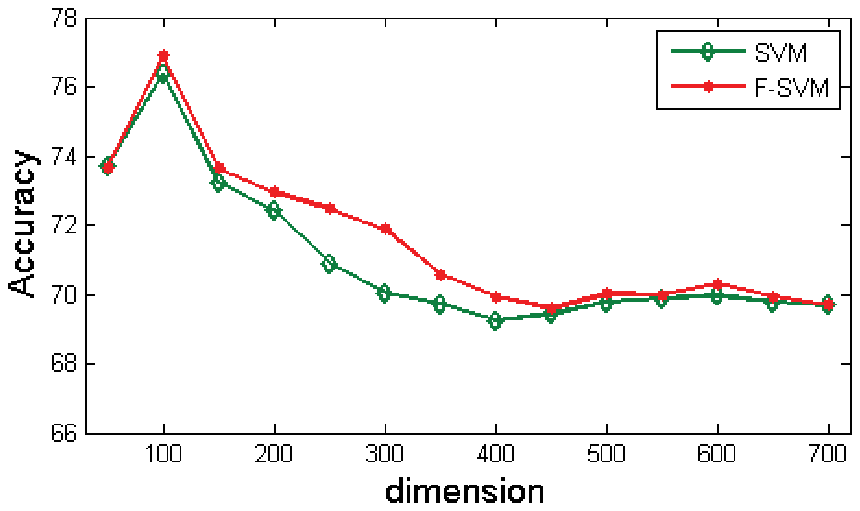}}

  \caption{ Classification accuracy (\%) of kernel SVM and kernel F-SVM under different kernel PCA dimensions. }
  \label{fig:fig2}
\end{figure}

In this subsection, we evaluate the performance of kernel F-SVM on the 11 UCI datasets. The Gaussian RBF kernel is adopted in our experiments, which includes an extra kernel parameter ${\sigma}$. As discussed in Section 4, we also consider the number of kernel PCA components in kernel F-SVM. Using four datasets, i.e., {\it Breast cancer}, {\it Thyroid}, {\it Heart}, and {\it German}, Fig.~\ref{fig:fig2} illustrates the classification accuracy of kernel SVM and kernel F-SVM under different kernel PCA dimensions. It is interesting to note that, the proper decreasing of kernel PCA dimension can consistently improve the classification accuracy both for kernel SVM and kernel F-SVM. Also from Fig.~\ref{fig:fig2} one can see that the kernel F-SVM is superior to kernel SVM under different dimensions. One possible explanation may be that the decreasing of kernel PCA dimension would make the learned transformation more stable.

Tables \ref{table4} and \ref{table5} list the mean classification accuracy and training time of five linear classifiers, i.e., kernel SVM, kernel RMM \cite{shivaswamy2010maximum}, kernel R-SVM$^{+}$ \cite{do2013convex}, kernel R-SVM$_{\mu}^{+}$ \cite{do2013convex}, and kernel F-SVM. For kernel methods, the superiority of F-SVM against the competing methods is more significant. The Kernel F-SVM outperforms kernel SVM on 10, kernel RMM \cite{shivaswamy2010maximum} on 9, kernel R-SVM$^{+}$ \cite{do2013convex} on 11, and kernel R-SVM$_{\mu}^{+}$ \cite{do2013convex} on 11 of all the 11 datasets in terms of classification accuracy. By training time, the kernel F-SVM is a little slower than SVM, but is about $10^3\!\!\sim\!\!10^4$ times faster than the other competing methods.

\subsection{Results on the LFW Database}
In this subsection, the LFW database is used to evaluate F-SVM for face verification. The database consists of more than 13,233 face images from 5,749 persons. The face images in the LFW database were collected from the Internet, and vary in pose, illumination, expression, and age, making LFW very suitable for studying unconstrained face verification. The face recognition method can be evaluated with two test protocols for LFW: the restricted and the unrestricted settings. Under the restricted setting, the only available information is whether each pair of training images is matched or not, and the performance of the face verification method is evaluated by 10-fold cross validation on a set of 3000 positive and 3000 negative image pairs.

\begin{figure}
  \centering
   \subfigure[Similar pairs]{
     \label{fig:subfig:a}
     \includegraphics[width=3.0in]{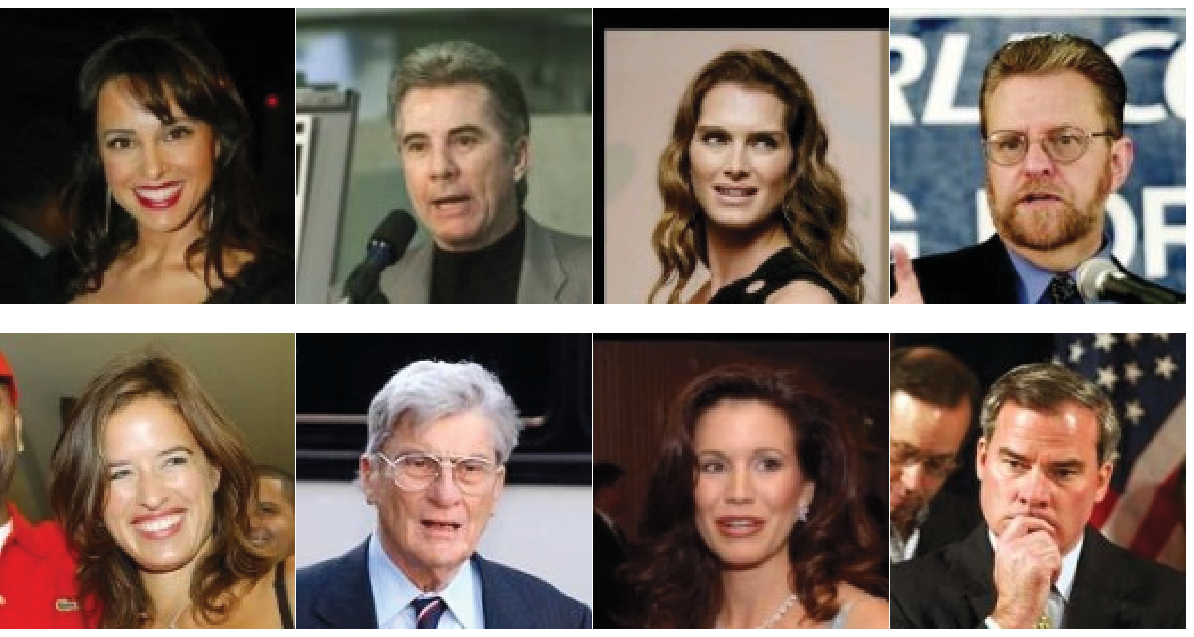}}

  \hspace{1in}
  \vspace{-0.1in}

   \subfigure[Dissimilar pairs]{
     \label{fig:subfig:b} 
     \includegraphics[width=3.0in]{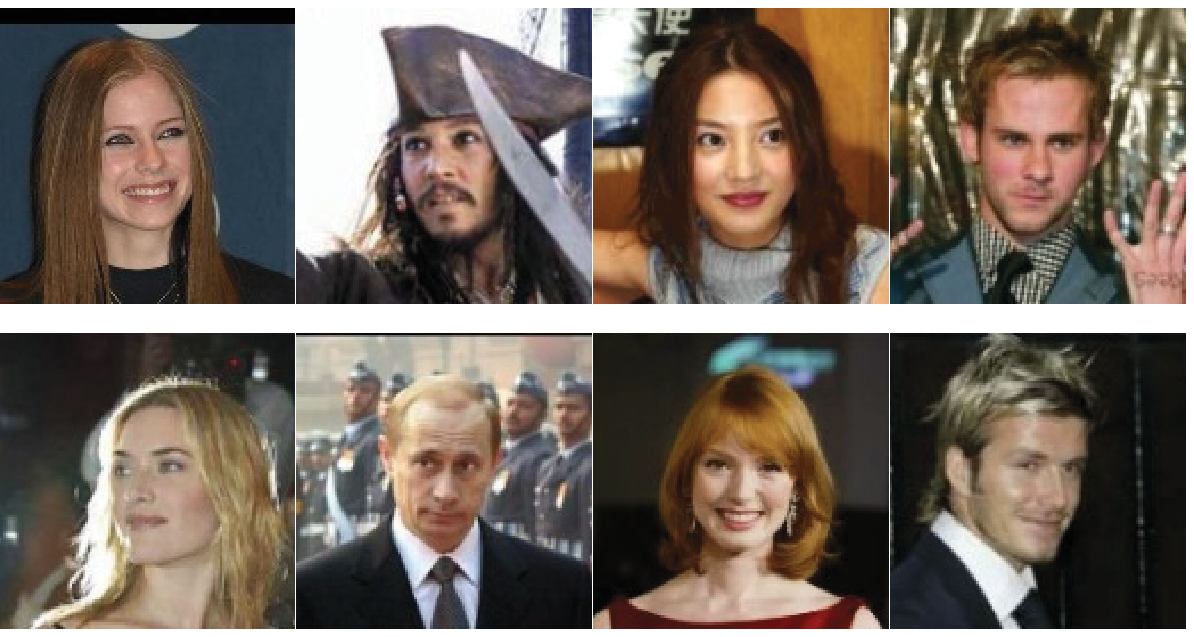}}

  \caption{ Examples of face image pairs in the LFW database.}
  \label{fig:fig3}
\end{figure}

In our experiment, we adopt the restricted setting with the face images aligned by the funneling method \cite{huang2007unsupervised}. Fig.~\ref{fig:fig3} shows some examples of similar and dissimilar pairs. We extract two kinds of features for each face image: SIFT feature and attribute feature, and compare F-SVM with SVM, RMM \cite{shivaswamy2010maximum}, R-SVM$^{+}$ \cite{do2013convex}, and R-SVM$_{\mu}^{+}$ \cite{do2013convex}, and several representative face verification methods, including LDML \cite{guillaumin2009you}, Nowak \cite{nowak2007learning}, V1-like/MKL \cite{pinto2009far}, and MERL+Nowak \cite{huang2008lfw}.

\begin{figure}
  \centering
   \subfigure[The attribute features]{
     \label{fig:subfig:a}
     \includegraphics[width=3.0in]{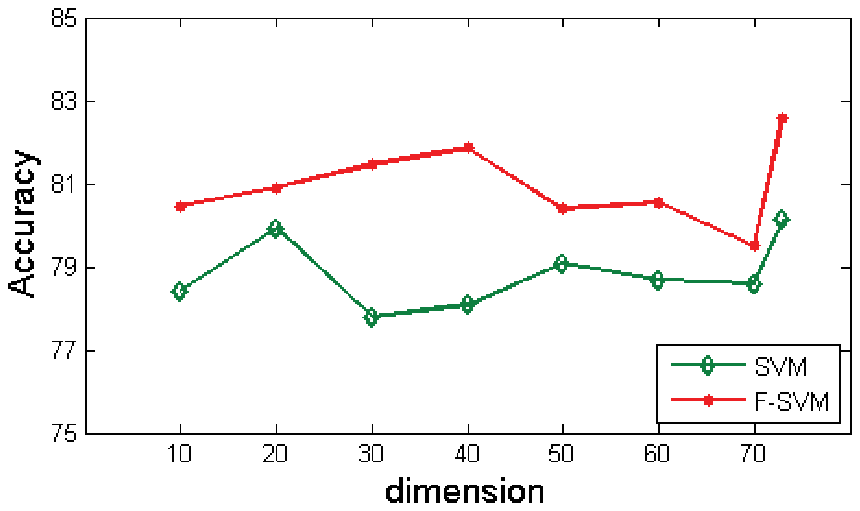}}

  \hspace{1in}
  \vspace{-0.1in}

   \subfigure[The combined features of SIFT and attributes]{
     \label{fig:subfig:b} 
     \includegraphics[width=3.0in]{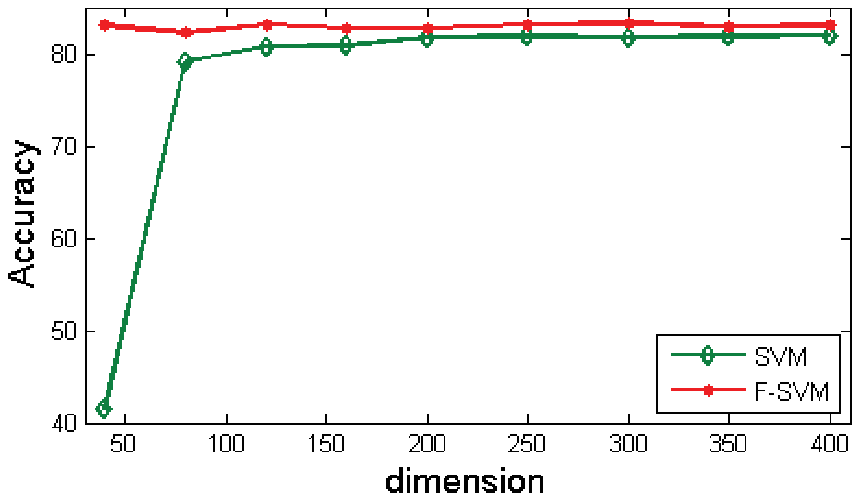}}

  \caption{ Examples of face image pairs in the LFW database.}
  \label{fig:fig4}
\end{figure}

Fig.~\ref{fig:fig4} shows the verification accuracies of SVM and F-SVM under different PCA dimensions by using the attribute feature and the combined features of SIFT and attributes, respectively. F-SVM using the combined features of SIFT and attributes (F-SVM-combined) achieves its best performance of 83.25\% when the dimension d = 300, and 82.58\% when the dimension d = 73 using the attribute features (F-SVM-attribute). SVM using the combined features of SIFT and attributes (SVM-combined) achieves its best performance of 81.90\% when the dimension d = 400, and 80.12\% when the dimension d = 73 using the attribute features (SVM-attribute). Thus, F-SVM can get better accuracy than SVM on the LFW database.

\begin{figure}
  \centering
   \subfigure[ROC curves]{
     \label{fig:subfig:a}
     \includegraphics[width=2.8in]{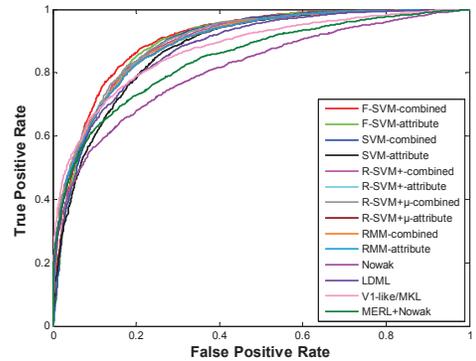}}

  \hspace{1in}
  \vspace{-0.1in}

   \subfigure[Cropped and zoom-in region of (a)]{
     \label{fig:subfig:b} 
     \includegraphics[width=2.8in]{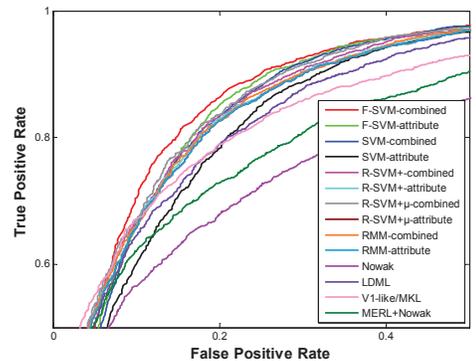}}

  \caption{ ROC curves of different face verification methods on the restricted LFW-funneled database.}
  \label{fig:fig5}
\end{figure}

We further compare F-SVM with several other face verification methods. Table 6 lists the accuracy of F-SVM, SVM, RMM \cite{shivaswamy2010maximum}, R-SVM$^{+}$ \cite{do2013convex}, R-SVM$_{\mu}^{+}$ \cite{do2013convex}, LDML \cite{guillaumin2009you},Nowak \cite{nowak2007learning}, V1-like/MKL \cite{pinto2009far}, and MERL+Nowak \cite{huang2008lfw}. We report the accuracy of F-SVM, SVM, RMM \cite{shivaswamy2010maximum}, R-SVM$^{+}$ \cite{do2013convex}, R-SVM$_{\mu}^{+}$ \cite{do2013convex}, LDML \cite{guillaumin2009you} using the attribute and the SIFT features, report the accuracy of Nowak \cite{nowak2007learning} and MERL+Nowak \cite{huang2008lfw} using SIFT and geometry feature, and report the accuracy of V1-like/MKL \cite{pinto2009far} using the V1-like features. For either the combined features of SIFT and attributes or the attribute features, F-SVM achieves higher accuracy than SVM, RMM \cite{shivaswamy2010maximum},R-SVM$^{+}$ \cite{do2013convex}, R-SVM$_{\mu}^{+}$ \cite{do2013convex}, LDML \cite{guillaumin2009you} separately from Table \ref{table6}. Fig.~\ref{fig:fig5} shows the ROC curves of the competing methods. Also one can see that F-SVM-combined gets better performance than other face verification methods.

\begin{table}[!t] \scriptsize
\renewcommand{\arraystretch}{1}
\caption{Comparison of accuracy obtained using different face verification methods. The top two results are shown in red and blue fonts, respectively.}
\label{table6} \centering
\begin{tabular}{|c|c|c|c|c|c|}
\hline
 Method, restricted    & Accuracy \\
\hline
F-SVM-combined           & \textcolor[rgb]{1.00,0.00,0.00}{$83.25$}    \\
\hline
F-SVM-attribute   & \textcolor[rgb]{0.00,0.07,1.00}{$82.58$}   \\
\hline
R-SVM$_{\mu}^{+}$-combined \cite{do2013convex}             &  $ 81.90 $ \\
\hline
R-SVM$_{\mu}^{+}$-attribute \cite{do2013convex}           & $ 81.50 $  \\
\hline
R-SVM$^{+}$-combined \cite{do2013convex}                   & $81.67$  \\
\hline
R-SVM$^{+}$-attribute \cite{do2013convex}                   &  $81.46$ \\
\hline
RMM-combined \cite{shivaswamy2010maximum}                & $81.30$ \\
\hline
RMM-attribute \cite{shivaswamy2010maximum}                  & $81.27$ \\
\hline
SVM-combined                 & $81.90$ \\
\hline
SVM-attribute                 & $80.12$ \\
\hline
Nowak \cite{nowak2007learning}                &  $73.93$ \\
\hline
LDML \cite{guillaumin2009you}  & $79.27$ \\
\hline
V1-like/MKL \cite{pinto2009far}  & $79.35$ \\
\hline
MERL+Nowak \cite{huang2008lfw}  & $76.18$ \\
\hline
\end{tabular}
\end{table}

\section{Conclusion}
In this paper, we proposed a convex radius-margin based SVM model (F-SVM) for joint learning of feature transformation and SVM classifier. For the formulation of F-SVM, lower and upper bounds of the radius of MEB are introduced to derive a novel approximation of radius-margin ratio, and all the individual inequality constraints are combined into one integrated inequality constraint, resulting in a convex relaxation of the radius-margin based SVM model. For model optimization, a semi-whitened PCA based method is proposed for the initialization of the learned transformation, and an alternating minimization algorithm is adopted to learn the feature transformation and SVM classifier. Further, F-SVM is kernelized by using kernel PCA. Our experimental results show that, F-SVM obtains higher classification accuracy than SVM and the state-of-the-art radius-margin based SVM methods, and is more efficient in training than the other radius-margin based SVM methods, e.g., RMM \cite{shivaswamy2010maximum}, R-SVM$^{+}$ \cite{do2013convex} and R-SVM$_{\mu}^{+}$ \cite{do2013convex}. In our future work, we will extend the proposed relaxed radius-margin based error bound to other classification methods and extend the proposed model for learning other forms of feature transformation tailored for specific applications.

%

\appendices
\section{}
{\textbf {Lemma A.1}}. $\bar{R}\ge R$.
\begin{proof}
Based on the definition of the radius, we get
\begin{eqnarray} \label{eq:class_A.1.1}
\begin{aligned}
 \nonumber {{R}^{2}} & =\min \limits_{{{\mathbf{x}}_{0}}} \max \limits_{i}\,{\left\| \mathbf{A}{{\mathbf{x}}_{i}}-\mathbf{A}{{\mathbf{x}}_{0}} \right\|_{2}^{2}} \\
 & \le \max \limits_{i}{\left\| \mathbf{A}{{\mathbf{x}}_{i}}-\mathbf{A\bar{x}} \right\|_{2}^{2}} \\
 & ={{{\bar{R}}}^{2}}.
\end{aligned}
\end{eqnarray}
\end{proof}

Denote ${{R}_{p}}$ by the maximum pairwise distance. We have
\begin{eqnarray} \label{eq:class_A.1.2}
 \nonumber {{R}_{p}}=\max \limits_{i,j}\{\left\| \mathbf{A}{{\mathbf{x}}_{i}}-\mathbf{A}{{\mathbf{x}}_{j}} \right\|_{2}^{2}\}.
\end{eqnarray}

{\textbf {Lemma A.2}} \cite{do2013convex}. $R\ge {{R}_{p}}/2$.

{\textbf {Lemma A.3}}. $\bar{R}\le {{R}_{p}}$.
\begin{proof}
Let ${{\mathbf{{x}'}}_{i}}=\mathbf{A}{{\mathbf{x}}_{i}}-\mathbf{A\bar{x}}$. We have $\mathbf{A}{{\mathbf{x}}_{i}}-\mathbf{A}{{\mathbf{x}}_{j}}={{\mathbf{{x}'}}_{i}}-{{\mathbf{{x}'}}_{j}}$. Based on the definition of $\bar{R}$
\begin{eqnarray} \label{eq:class_A.3.1}
 \nonumber {{\bar{R}}^{2}}=\max \limits_{i} \left\{ {{\left\| {{{\mathbf{{x}'}}}_{i}} \right\|}^{2}} \right\}={{\left\| {{{\mathbf{{x}'}}}_{{{i}^{*}}}} \right\|}^{2}},
\end{eqnarray}
we will prove that there exist some ${{j}^{*}}$ which makes ${{\left\| {{{\mathbf{{x}'}}}_{{{i}^{*}}}}-{{{\mathbf{{x}'}}}_{{{j}^{*}}}} \right\|}^{2}}\ge {{\left\| {{{\mathbf{{x}'}}}_{{{i}^{*}}}} \right\|}^{2}}$. Based on the definition of $\mathbf{\bar{x}}$, we have $\sum\nolimits_{j}{{{{\mathbf{{x}'}}}_{j}}}=0$. Then, we derive
\begin{eqnarray} \label{eq:class_A.3.2}
 \nonumber \sum\nolimits_{j}{{{{\mathbf{{x}'}}}_{j}}{{{\mathbf{{x}'}}}_{{{i}^{*}}}}}=0\Rightarrow \min \limits_{j}\left\{ {{{\mathbf{{x}'}}}_{j}}{{{\mathbf{{x}'}}}_{{{i}^{*}}}} \right\}={{\mathbf{{x}'}}_{{{j}^{*}}}}{{\mathbf{{x}'}}_{{{i}^{*}}}}\le 0.
\end{eqnarray}
Since ${{\left\| {{{\mathbf{{x}'}}}_{{{j}^{*}}}} \right\|}^{2}}\ge 0$ and $-2{{\mathbf{{x}'}}_{{{j}^{*}}}}{{\mathbf{{x}'}}_{{{i}^{*}}}}\ge 0$, one can easily see that
\begin{eqnarray} \label{eq:class_A.3.3}
 \nonumber {{\left\| {{{\mathbf{{x}'}}}_{{{i}^{*}}}}-{{{\mathbf{{x}'}}}_{{{j}^{*}}}} \right\|}^{2}}\ge {{\left\| {{{\mathbf{{x}'}}}_{{{i}^{*}}}} \right\|}^{2}}.
\end{eqnarray}
Based on the definition of ${{R}_{p}}$:
\begin{eqnarray} \label{eq:class_A.3.4}
 \nonumber R_{p}^{2}\ge {{\left\| {{{\mathbf{{x}'}}}_{{{i}^{*}}}}-{{{\mathbf{{x}'}}}_{{{j}^{*}}}} \right\|}^{2}}.
\end{eqnarray}

Combining the two inequalities above, we can prove $\bar{R}\le {{R}_{p}}$.
\end{proof}

Finally, by combining {\textbf {Lemmas}} 1$\sim$3, we obtain the following theorem:

{\textbf {Theorem A.1}}. The margin R is bounded by $\bar{R}$ by:
\begin{eqnarray} \label{eq:class_Theorem A.1}
\nonumber \frac{1}{2}\overline{R}\le R\le \overline{R}.
\end{eqnarray}

\section{}
{\textbf {Lemma B.1}} \cite{petersen2008matrix}. Given two symmetric positive definite (SPD) matrices ${\bf A}$ and ${\bf B}$, we have
\begin{eqnarray} \label{eq:class_28}
{{\left( \mathbf{A}+\mathbf{B} \right)}^{-1}}={{\mathbf{A}}^{-1}}-{{\mathbf{A}}^{-1}}{{({{\mathbf{A}}^{-1}}+{{\mathbf{B}}^{-1}})}^{-1}}{{\mathbf{A}}^{-1}},
\end{eqnarray}
\begin{eqnarray} \label{eq:class_29}
\begin{aligned}
{{\left( \mathbf{A}+\mathbf{B}\right)}^{-1}} & = {{\mathbf{A}}^{-1}}{{({{\mathbf{A}}^{-1}}+{{\mathbf{B}}^{-1}})}^{-1}}{{\mathbf{B}}^{-1}} \\
& = {{\mathbf{B}}^{-1}}{{({{\mathbf{A}}^{-1}}+{{\mathbf{B}}^{-1}})}^{-1}}{{\mathbf{A}}^{-1}}.
\end{aligned}
\end{eqnarray}

{\textbf {Theorem B.2}}. The problem Eq. (\ref{eq:class_F-SVM-2}) is a convex optimization problem.
\begin{proof}
Note that all the constraints define a convex set, and $\sum\nolimits_{i}{{{\xi }_{i}}}$ and $tr\left( \mathbf{MS} \right)$ are linear to $\mathbf{\xi }$ and ${\bf M}$, respectively. Then the key step is to prove that the function ${{\mathbf{w}}^{T}}{{\mathbf{M}}^{-1}}\mathbf{w}$ is convex for $\mathbf{M}\succ 0$, i.e., for any $1\ge \theta \ge 0$,
\begin{eqnarray} \label{eq:class_B.2.1}
\begin{aligned}
\nonumber & \theta \mathbf{w}_{1}^{T}\mathbf{M}_{1}^{-1}{{\mathbf{w}}_{1}}+(1-\theta )\mathbf{w}_{2}^{T}\mathbf{M}_{2}^{-1}{{\mathbf{w}}_{2}} \ge \\
 & {{\left( \theta {{\mathbf{w}}_{1}}+(1-\theta ){{\mathbf{w}}_{2}} \right)}^{T}}{{\left( \theta {{\mathbf{M}}_{1}}+(1-\theta ){{\mathbf{M}}_{2}} \right)}^{-1}}\left( \theta {{\mathbf{w}}_{1}}+(1-\theta ){{\mathbf{w}}_{2}} \right).
\end{aligned}
\end{eqnarray}
Note that ${{\left( \theta {{\mathbf{w}}_{1}}\!\!+\!\!(1\!\!-\!\!\theta ){{\mathbf{w}}_{2}} \right)}^{T}}{{\left( \theta {{\mathbf{M}}_{1}}\!\!+\!\!(1\!\!-\!\!\theta ){{\mathbf{M}}_{2}} \right)}^{-1}}\left( \theta {{\mathbf{w}}_{1}}\!\!+\!\!(1\!\!-\!\!\theta ){{\mathbf{w}}_{2}} \right)$ contains three terms:
\begin{eqnarray} \label{eq:class_B.2.2}
\nonumber {{\theta }^{2}}\mathbf{w}_{1}^{T}{{\left( \theta {{\mathbf{M}}_{1}}+(1-\theta ){{\mathbf{M}}_{2}} \right)}^{-1}}{{\mathbf{w}}_{1}} \\
\nonumber {{(1-\theta )}^{2}}\mathbf{w}_{2}^{T}{{\left( \theta {{\mathbf{M}}_{1}}+(1-\theta ){{\mathbf{M}}_{2}} \right)}^{-1}}{{\mathbf{w}}_{2}} \\
\nonumber \theta (1-\theta )\mathbf{w}_{1}^{T}{{\left( \theta {{\mathbf{M}}_{1}}+(1-\theta ){{\mathbf{M}}_{2}} \right)}^{-1}}{{\mathbf{w}}_{2}}.
\end{eqnarray}
First, we have
\begin{eqnarray} \label{eq:class_30}
\begin{aligned}
  & {{\theta }^{2}}{{\left( \theta {{\mathbf{M}}_{1}}+(1-\theta ){{\mathbf{M}}_{2}} \right)}^{-1}}=\theta {{\left( {{\mathbf{M}}_{1}}+\frac{(1-\theta )}{\theta }{{\mathbf{M}}_{2}} \right)}^{-1}} \\
  &=\theta \!\! \left( \!\! \mathbf{M}_{1}^{-1}\!\!-\!\!\mathbf{M}_{1}^{-1}{{\left( \mathbf{M}_{1}^{-1}\!\!+\!\!{{\left( \frac{(1\!\!-\!\!\theta )}{\theta }{{\mathbf{M}}_{2}} \right)}^{-1}} \right)}^{-1}}\mathbf{M}_{1}^{-1} \!\! \right) \\
  &=\theta \mathbf{M}_{1}^{-1}-\mathbf{M}_{1}^{-1}{{\left( {{\left( \theta {{\mathbf{M}}_{1}} \right)}^{-1}}+{{\left( (1-\theta ){{\mathbf{M}}_{2}} \right)}^{-1}} \right)}^{-1}}\mathbf{M}_{1}^{-1},
\end{aligned}
\end{eqnarray}
and then we get
\begin{eqnarray} \label{eq:class_31}
\begin{aligned}
 & \theta \mathbf{w}_{1}^{T}\mathbf{M}_{1}^{-1}{{\mathbf{w}}_{1}}-{{\theta }^{2}}\mathbf{w}_{1}^{T}{{\left( \theta {{\mathbf{M}}_{1}}+(1-\theta ){{\mathbf{M}}_{2}} \right)}^{-1}}{{\mathbf{w}}_{1}} \\
 & =\mathbf{w}_{1}^{T}\mathbf{M}_{1}^{-1}{{\left( {{\left( \theta {{\mathbf{M}}_{1}} \right)}^{-1}}+{{\left( (1-\theta ){{\mathbf{M}}_{2}} \right)}^{-1}} \right)}^{-1}}\mathbf{M}_{1}^{-1}{{\mathbf{w}}_{1}}.
\end{aligned}
\end{eqnarray}
Analogously, we get
\begin{eqnarray} \label{eq:class_32}
\begin{aligned}
  & (1\!-\! \theta )\mathbf{w}_{2}^{T}\mathbf{M}_{2}^{-1}{{\mathbf{w}}_{2}}\!-\!{{(1\!-\!\theta )}^{2}}\mathbf{w}_{2}^{T}{{\left( \theta {{\mathbf{M}}_{1}}\!+\!(1\!-\!\theta ){{\mathbf{M}}_{2}} \right)}^{-1}}{{\mathbf{w}}_{2}} \\
  & =\mathbf{w}_{2}^{T}\mathbf{M}_{2}^{-1}{{\left( {{\left( \theta {{\mathbf{M}}_{1}} \right)}^{-1}}+{{\left( (1\!-\!\theta ){{\mathbf{M}}_{2}} \right)}^{-1}} \right)}^{-1}}\mathbf{M}_{2}^{-1}{{\mathbf{w}}_{2}}.
\end{aligned}
\end{eqnarray}
With Eq. (\ref{eq:class_30}), we have
\begin{eqnarray} \label{eq:class_33}
\begin{aligned}
 & \mathbf{w}_{1}^{T}{{\left( \theta {{\mathbf{M}}_{1}}+(1-\theta ){{\mathbf{M}}_{2}} \right)}^{-1}}{{\mathbf{w}}_{2}} \\
 & =\mathbf{w}_{1}^{T}\mathbf{M}_{1}^{-1}{{\left( {{\left( \theta {{\mathbf{M}}_{1}} \right)}^{-1}}+{{\left( (1-\theta ){{\mathbf{M}}_{2}} \right)}^{-1}} \right)}^{-1}}\mathbf{M}_{2}^{-1}{{\mathbf{w}}_{2}}.
\end{aligned}
\end{eqnarray}
Combining Eqns. (29)-(32), we get
\begin{eqnarray} \label{eq:34}
\begin{aligned}
 \nonumber & \theta \mathbf{w}_{1}^{T}\mathbf{M}_{1}^{-1}{{\mathbf{w}}_{1}}\!+\!(1\!-\!\theta )\mathbf{w}_{2}^{T}\mathbf{M}_{2}^{-1}{{\mathbf{w}}_{2}} \\
 & \!-{{\left( \theta {{\mathbf{w}}_{1}}\!\!+\!\!(1\!-\!\theta ){{\mathbf{w}}_{2}} \right)}^{T}}{{\left( \theta {{\mathbf{M}}_{1}}\!\!+\!\!(1\!-\!\theta ){{\mathbf{M}}_{2}} \right)}^{\!-\!1}}\left( \theta {{\mathbf{w}}_{1}}\!\!+\!\!(1\!-\!\theta ){{\mathbf{w}}_{2}} \right) \\
 & \!= \mathbf{w}_{1}^{T}\mathbf{M}_{1}^{\!-\!1}{{\left( {{\left( \theta {{\mathbf{M}}_{1}} \right)}^{\!-\!1}}\!\!+\!\!{{\left( (1\!-\!\theta ){{\mathbf{M}}_{2}} \right)}^{\!-\!1}} \right)}^{\!-1\!}}\mathbf{M}_{1}^{\!-\!1}{{\mathbf{w}}_{1}}\\
 & \!+ \mathbf{w}_{2}^{T}\mathbf{M}_{2}^{\!-\!1}{{\left( {{\left( \theta {{\mathbf{M}}_{1}} \right)}^{\!-\!1}}\!\!+\!\!{{\left( (1\!-\!\theta ){{\mathbf{M}}_{2}} \right)}^{\!-\!1}} \right)}^{\!-1\!}}\mathbf{M}_{2}^{\!-\!1}{{\mathbf{w}}_{2}} \\
 & \!- \mathbf{w}_{1}^{T}\mathbf{M}_{1}^{\!-\!1}{{\left( {{\left( \theta {{\mathbf{M}}_{1}} \right)}^{\!-\!1}}\!\!+\!\!{{\left( (1\!-\!\theta ){{\mathbf{M}}_{2}} \right)}^{\!-\!1}} \right)}^{\!-1\!}}\mathbf{M}_{2}^{\!-\!1}{{\mathbf{w}}_{2}} \\
 & \!= {{\left\| {{\left( {{\left( \theta {{\mathbf{M}}_{1}} \right)}^{\!-\!1}}\!+\!{{\left( (1\!-\!\theta ){{\mathbf{M}}_{2}} \right)}^{\!-\!1}} \right)}^{-\frac{1}{2}}}\left( \mathbf{M}_{1}^{\!-\!1}{{\mathbf{w}}_{1}}\!-\!\mathbf{M}_{2}^{\!-\!1}{{\mathbf{w}}_{2}} \right) \right\|}^{2}} \\
 & \ge 0.
\end{aligned}
\end{eqnarray}
Thus, the F-SVM model is convex.
\end{proof}

\section{}
{\textbf {Theorem 4}}. Given a SPD matrix ${\bf S}$ and ${\tau }'>0$, $\mathbf{\hat{B}}$ defined in Eq. (\ref{eq:class_17}) is the optimal solution to the problem:
\begin{eqnarray} \label{eq:class_C.1}
 \nonumber \mathbf{\hat{B}}=\arg \min \limits_{\mathbf{B}}\left\{ L(\mathbf{B},{\tau }')={{\left\| \mathbf{B} \right\|}_{*}}+{\tau }'\left( tr\left( {{\mathbf{B}}^{-1}}\mathbf{S} \right) \right) \right\}.
\end{eqnarray}

\begin{proof}
$L(\mathbf{B},{\tau }')$ is strictly convex with respect to ${\bf B}$ [28]. Given $g(\mathbf{B})={\tau }'tr\left( {{\mathbf{B}}^{-1}}\mathbf{S} \right)$, we have:
\begin{eqnarray} \label{eq:class_C.2}
 \nonumber \frac{\partial g}{\partial \mathbf{B}}=-{\tau }'{{\left( {{\mathbf{B}}^{-1}}\mathbf{S}{{\mathbf{B}}^{-1}} \right)}^{T}}
\end{eqnarray}

From \cite{cai2010singular, watson1992characterization}, the set of sub-gradients of the nuclear norm $\partial {{\left\| \mathbf{B} \right\|}_{*}}$ can be represented as
\begin{eqnarray} \label{eq:class_C.3}
 \nonumber \partial {{\left\| \mathbf{B} \right\|}_{*}} \!\!=\!\! \{\mathbf{\bar{U}\bar{\Sigma }}{{\mathbf{\bar{U}}}^{T}}\!\!+\!\!\mathbf{W}|\mathbf{W} \!\! \in \!\! {{\mathbf{R}}^{d \!\times\! d}},{{\mathbf{\bar{U}}}^{T}}\mathbf{W}\!\!=\!\!0,\mathbf{W\bar{U}}\!\!=\!\!0,{{\left\| \mathbf{W} \right\|}_{2}}\!\!\le\!\! 1\},
\end{eqnarray}
where $\mathbf{\bar{U}\bar{\Sigma }}{{\mathbf{\bar{U}}}^{T}}$ is the eigenvalue decomposition of ${\bf B}$, each column of $\mathbf{\bar{U}}$ is a eigenvector, $\mathbf{\bar{\Sigma }}$ is a diagonal matrix with $\mathbf{\bar{\Sigma }}=diag({{\sigma }_{1}},{{\sigma }_{2}},\cdots ,{{\sigma }_{d}})$ ($0\le {{\sigma }_{1}}\le {{\sigma }_{2}}\le \cdots \le {{\sigma }_{d}}$).

To prove that $\mathbf{\hat{B}}$ is the optimal solution, we will show that
\begin{eqnarray} \label{eq:class_C.4}
 \nonumber 0\in -{\tau }'{{\left( {{{\mathbf{\hat{B}}}}^{-1}}\mathbf{S}{{{\mathbf{\hat{B}}}}^{-1}} \right)}^{T}}+\partial {{\left\| {\mathbf{\hat{B}}} \right\|}_{*}}.
\end{eqnarray}

With the matrix $\mathbf{\hat{B}}$ in Eq. (\ref{eq:class_B}), we have
\begin{eqnarray} \label{eq:class_C.4}
 \nonumber {\tau }'{{\left( {{{\mathbf{\hat{B}}}}^{-1}}\mathbf{S}{{{\mathbf{\hat{B}}}}^{-1}} \right)}^{T}}=\mathbf{U}{{\mathbf{U}}^{T}}
\end{eqnarray}

Let ${\bf W}=0$. We have ${{\mathbf{U}}^{T}}\mathbf{W}=0$, $\mathbf{WU}=0$, and ${{\left\| \mathbf{W} \right\|}_{2}}\le 1$. Thus $\mathbf{U}{{\mathbf{U}}^{T}}\in \partial {{\left\| {\mathbf{\hat{B}}} \right\|}_{*}}$, and $\mathbf{\hat{B}}$ is the optimal minimizer of $L(\mathbf{B},{\tau }')$.
\end{proof}

\section*{Acknowledgment}

The work is partially supported by the NSFC fund of China under contract NO. 61271093, the program of ministry of education for New Century Excellent Talents in University under Grant NCET-12-0150.

\ifCLASSOPTIONcaptionsoff
  \newpage
\fi



\bibliographystyle{IEEEtran}
\bibliography{IEEEabrv,fsvm}
%

%

\end{document}